\documentclass{baai}
\usepackage{float}
\usepackage{placeins}
\usepackage{siunitx}
\usepackage{tcolorbox}
\tcbuselibrary{skins}
\usepackage{multirow}
\usepackage{makecell}
\usepackage{pifont}
\usepackage{marvosym}
\usepackage{colortbl}
\newcommand{\correnvelope}{\scalebox{1.35}{\Letter}}
\usepackage{xspace}
\usepackage{dirtree}
\definecolor{qboxbg}{HTML}{F7F8FA}
\definecolor{qteal}{HTML}{0E8F8F}
\definecolor{qink}{HTML}{374151}
\definecolor{thmframe}{HTML}{BBBFC7}
\definecolor{thmbg}{HTML}{FAFAFA}
\usepackage[nameinlink,noabbrev]{cleveref}
\crefname{section}{Section}{Sections}
\Crefname{section}{Section}{Sections}
\crefname{table}{Table}{Tables}
\Crefname{table}{Table}{Tables}
\crefname{figure}{Figure}{Figures}
\Crefname{figure}{Figure}{Figures}
\crefname{equation}{Equation}{Equations}
\Crefname{equation}{Equation}{Equations}
\setcitestyle{numbers,square,comma}
\definecolor{tierblue}{RGB}{28, 95, 140}
\definecolor{tierteal}{RGB}{15, 107, 99}
\definecolor{tierviolet}{RGB}{91, 69, 168}
\definecolor{linkred}{RGB}{200, 30, 30}
\definecolor{okgreen}{RGB}{28, 128, 96}
\definecolor{lgreen}{RGB}{231, 242, 232}
\definecolor{tiertxt}{gray}{0.40}
\newcommand{\pc}[1]{\textcolor{tiertxt}{#1}}

\newcommand{\dataset}{HumanCOIN}

\newcommand{\rawHoursLone}{10000}
\newcommand{\rawHoursLtwo}{2000}
\newcommand{\rawHoursLthree}{500}
\newcommand{\zoneResidential}{9}
\newcommand{\zoneCommercial}{10}
\newcommand{\zoneProfessional}{5}
\newcommand{\zoneIndustrial}{6}
\newcommand{\zonePublic}{7}
\newcommand{\numzones}{5}
\newcommand{\zoneResidual}{1}
\newcommand{\rawScenesLthree}{9}
\newcommand{\hoursResidential}{\num{4521}}
\newcommand{\hoursCommercial}{\num{2588}}
\newcommand{\hoursProfessional}{\num{1982}}
\newcommand{\hoursIndustrial}{\num{1957}}
\newcommand{\hoursPublic}{\num{1211}}
\newcommand{\hoursResidual}{\num{242}}
\newcommand{\hoursKitchen}{\num{1175}}
\newcommand{\hoursFactory}{\num{1168}}
\newcommand{\hoursOffice}{\num{1129}}
\newcommand{\hoursBalcony}{\num{13}}
\newcommand{\pctScenesLthreeOfLone}{33.2}
\newcommand{\pctScenesLthreeOfLtwo}{50.9}
\newcommand{\numscenesLtwo}{21}
\newcommand{\pctScenesLtwo}{55}
\newcommand{\numsubverbs}{99}
\newcommand{\nummotionverbs}{46}
\newcommand{\numsubverbsLthree}{40}
\newcommand{\nummotionverbsLthree}{25}
\newcommand{\pctSubverbsTop}{60.0}
\newcommand{\pctMotionverbsTop}{92.5}
\newcommand{\medSubverbClass}{81}
\newcommand{\pctSubverbsLthreeOfLone}{77.3}
\newcommand{\pctMotionverbsLthreeOfLone}{95.2}
\newcommand{\rawObjects}{635}
\newcommand{\rawObjectsLone}{630}
\newcommand{\rawObjectsLtwo}{386}
\newcommand{\rawObjectsLthree}{95}
\newcommand{\pctObjectsLthreeOfLone}{82.0}
\newcommand{\pctObjectsLthreeOfLtwo}{97.3}
\newcommand{\pctObjectsTop}{42.5}
\newcommand{\medObjectClass}{130}
\newcommand{\secPeak}{52}
\newcommand{\secMean}{88}
\newcommand{\secMeanLone}{101}
\newcommand{\secMeanLtwo}{65}
\newcommand{\secMeanLthree}{60}
\newcommand{\secShort}{10}
\newcommand{\secTail}{5}
\newcommand{\subMeanLone}{11.0}
\newcommand{\subMeanLtwo}{8.3}
\newcommand{\subMeanLthree}{7.9}
\newcommand{\numsites}{6}
\newcommand{\annotTotal}{1.35B}
\newcommand{\annotLone}{1.08B}
\newcommand{\annotLtwo}{216M}
\newcommand{\annotLthree}{54M}

\newcommand{\hoursLone}{\num{\rawHoursLone}}
\newcommand{\hoursLtwo}{\num{\rawHoursLtwo}}
\newcommand{\hoursLthree}{\num{\rawHoursLthree}}
\newcommand{\numhours}{\num{\fpeval{\rawHoursLone+\rawHoursLtwo+\rawHoursLthree}}}
\newcommand{\shareLone}{\fpeval{round(100*\rawHoursLone/(\rawHoursLone+\rawHoursLtwo+\rawHoursLthree),0)}}
\newcommand{\shareLtwo}{\fpeval{round(100*\rawHoursLtwo/(\rawHoursLone+\rawHoursLtwo+\rawHoursLthree),0)}}
\newcommand{\shareLthree}{\fpeval{round(100*\rawHoursLthree/(\rawHoursLone+\rawHoursLtwo+\rawHoursLthree),0)}}
\newcommand{\numscenesZoned}{\fpeval{\zoneResidential+\zoneProfessional+\zoneIndustrial+\zoneCommercial+\zonePublic}}
\newcommand{\numscenes}{\fpeval{\numscenesZoned+\zoneResidual}}
\newcommand{\numscenesLthree}{\rawScenesLthree}
\newcommand{\numscenesAbsent}{\fpeval{\numscenesZoned+\zoneResidual-\rawScenesLthree}}
\newcommand{\pctScenesLthree}{\fpeval{round(100*\rawScenesLthree/(\numscenesZoned+\zoneResidual),0)}}
\newcommand{\numobjects}{\num{\rawObjects}}
\newcommand{\numobjectsAbsent}{\num{\fpeval{\rawObjects-\rawObjectsLthree}}}
\newcommand{\pctObjectsKept}{\fpeval{round(100*\rawObjectsLthree/\rawObjects,0)}}

\newcommand{\tbd}{\textcolor{red}{\textbf{TBD}}\xspace}
\title{HumanCOIN: A 12,500-Hour Dataset for Human Intent in Egocentric Manipulation}
\newcommand{\authormetadata}{%
  Shihan Wu, You Liu, Chunyu He, Minghui Qin, Longhan Jiao, Jinbang Guo, Shaoxuan Xie,
  Kaixi Tian, Wenjie Chen, Jingrui Pang, Yuanhang Zhang, Yichi Zhang, Miao Liu, Tengfei Pan,
  Yance Jiao, Yunyan Zhang, Yihang Li, Yongjian Shen, Li Liu, Kai Zhu, Chen Fang,
  Jianxing Chen, Sihan Liu, Xiaofeng Wang, Jie Zhao, Jing Li, Yueqi Wang, Yichen Tang,
  Wei Li, Zhexuan Yang, Yue Cao, Ji Zhao, Kai Wang, Dongsheng Jiang, Yongwei Wang, Wei Mu,
  Jiangmiao Pang, Maoqing Yao, Wenchao Ding, Zheng Zhu, Zhongpu Xia, Dalong Du, Jianwei Niu,
  Dahui Gao, Zhiwei Yu, Bo Lei, Huan Yang, Yonghua Lin, Tiejun Huang, Zhongyuan Wang,
  Guocai Yao%
}
\author{\authormetadata}
\reportauthors{%
  Shihan~Wu\textsuperscript{1} \enspace 
  You~Liu\textsuperscript{1} \enspace 
  Chunyu~He\textsuperscript{2,1} \enspace 
  Minghui~Qin\textsuperscript{1} \enspace 
  Longhan~Jiao\textsuperscript{3,1} \enspace 
  Jinbang~Guo\textsuperscript{1} \enspace 
  Shaoxuan~Xie\textsuperscript{1} \\
  Kaixi~Tian\textsuperscript{1} \enspace 
  Wenjie~Chen\textsuperscript{1} \enspace 
  Jingrui~Pang\textsuperscript{4,1} \enspace 
  Yuanhang~Zhang\textsuperscript{1} \enspace 
  Yichi~Zhang\textsuperscript{5} \enspace 
  Miao~Liu\textsuperscript{5} \\
  Tengfei~Pan\textsuperscript{1} \enspace 
  Yance~Jiao\textsuperscript{1} \enspace 
  Yunyan~Zhang\textsuperscript{1} \enspace 
  Yihang~Li\textsuperscript{6} \enspace 
  Yongjian~Shen\textsuperscript{7} \enspace 
  Li~Liu\textsuperscript{7} \\
  Kai~Zhu\textsuperscript{8} \enspace 
  Chen~Fang\textsuperscript{9} \enspace 
  Jianxing~Chen\textsuperscript{10} \enspace 
  Sihan~Liu\textsuperscript{10} \enspace 
  Xiaofeng~Wang\textsuperscript{11} \enspace 
  Jie~Zhao\textsuperscript{12} \\
  Jing~Li\textsuperscript{13} \enspace 
  Yueqi~Wang\textsuperscript{14} \enspace 
  Yichen~Tang\textsuperscript{15} \enspace 
  Wei~Li\textsuperscript{16} \enspace 
  Zhexuan~Yang\textsuperscript{17} \enspace 
  Yue~Cao\textsuperscript{18} \\
  Ji~Zhao\textsuperscript{19} \enspace 
  Kai~Wang\textsuperscript{20} \enspace 
  Dongsheng~Jiang\textsuperscript{21} \enspace 
  Yongwei~Wang\textsuperscript{22} \enspace 
  Wei~Mu\textsuperscript{23} \enspace 
  Jiangmiao~Pang\textsuperscript{6} \\
  Maoqing~Yao\textsuperscript{7} \enspace 
  Wenchao~Ding\textsuperscript{24,25} \enspace 
  Zheng~Zhu\textsuperscript{11} \enspace 
  Zhongpu~Xia\textsuperscript{13} \enspace 
  Dalong~Du\textsuperscript{26} \enspace 
  Jianwei~Niu\textsuperscript{27} \\
  Dahui~Gao\textsuperscript{1} \enspace 
  Zhiwei~Yu\textsuperscript{1} \enspace 
  Bo~Lei\textsuperscript{1} \enspace 
  Huan~Yang\textsuperscript{1} \enspace 
  Yonghua~Lin\textsuperscript{1} \enspace 
  Tiejun~Huang\textsuperscript{2,1} \\
  Zhongyuan~Wang\textsuperscript{1 \correnvelope} \enspace 
  Guocai~Yao\textsuperscript{1 \correnvelope}%
}
\newcommand{\aff}[2]{\textsuperscript{#1}\,#2}
\reportaffiliation{{
  \aff{1}{Beijing Academy of Artificial Intelligence} \enspace
  \aff{2}{Peking University} \enspace
  \aff{3}{University of Science and Technology of China} \\
  \aff{4}{Kraxis AI} \enspace
  \aff{5}{College of AI, Tsinghua University} \enspace
  \aff{6}{JD Cloud} \enspace
  \aff{7}{Maniformer} \enspace
  \aff{8}{Ant Digital Technologies} \\
  \aff{9}{Lightwheel AI} \enspace
  \aff{10}{GenRobot AI} \enspace
  \aff{11}{GigaAI} \enspace
  \aff{12}{BODEN AI} \enspace
  \aff{13}{Anyverse Dynamics} \enspace
  \aff{14}{ATTA Labs} \enspace
  \aff{15}{Gentelli Robots} \\
  \aff{16}{HuaZhi.AI} \enspace
  \aff{17}{ArcheBase} \enspace
  \aff{18}{IO-AI Tech} \enspace
  \aff{19}{Magine AI} \enspace
  \aff{20}{DamonGroup} \enspace
  \aff{21}{LuMiWorld} \enspace
  \aff{22}{PRT} \enspace
  \aff{23}{Synapath} \\
  \aff{24}{TARS} \enspace
  \aff{25}{Fudan University} \enspace
  \aff{26}{SynapX} \enspace
  \aff{27}{School of Computer Science and Engineering, Beihang University}\par}}
\reportauthornote{\textsuperscript{\Letter}\,Corresponding authors: Zhongyuan Wang and
  Guocai Yao (\href{mailto:gcyao1@baai.ac.cn}{gcyao1@baai.ac.cn}); contact:
  \href{mailto:shihan.wu.koorye@outlook.com}{shihan.wu.koorye@outlook.com}}
\reportlogo{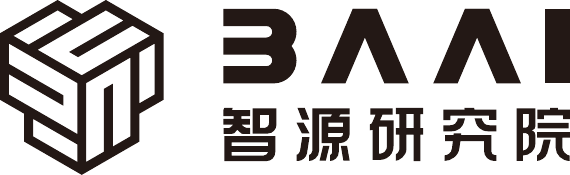}
\reportgroup{BAAI -- AhaData Group}
\reportdate{November 2026}
\reportcovertitlegap{14pt}
\reportcoverauthorfont{\bfseries\fontsize{10}{13}\selectfont}
\reportcoveraffiliationfont{\normalfont\fontsize{9}{12}\selectfont}
\reportcoverabstractgap{20pt}
\reportabstractsize{\fontsize{9}{10.7}\selectfont}
\reportabstractindent{1.4em}
\reportcoverimage[0.94\linewidth]{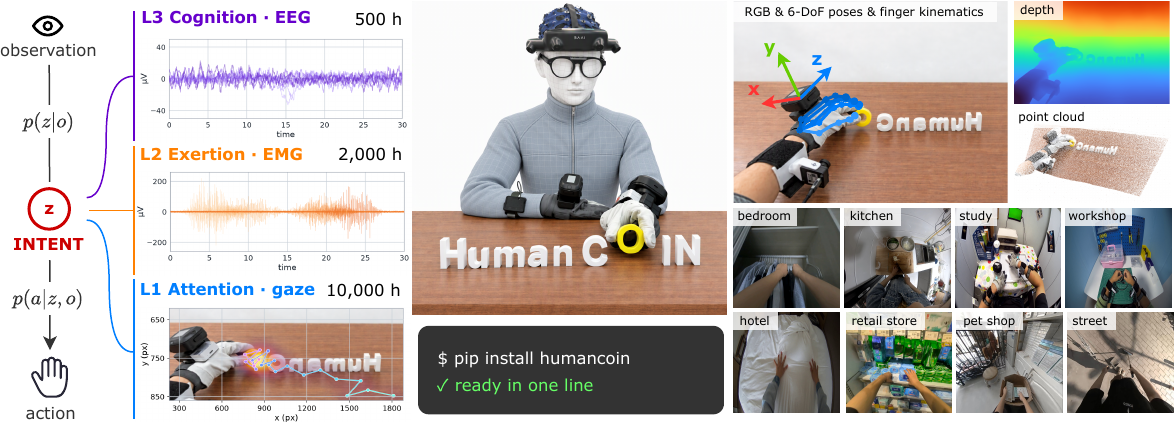}
\reportabstract{First-person manipulation datasets treat the observation--action pair as the
atomic unit and compress the generation and execution of an action into a policy
network's weights, rendering the intent state behind the action unobservable.
Human motor control research shows that intent mediates between observation and
action and leaves observable traces along the \emph{intent pathway} from decision
to effect: gaze reaches the target before the hand, muscle load builds before
contact, and cortical activity precedes the motor command.
To record this pathway at scale, we present \dataset{}, a first-person
manipulation dataset built around human intent: \numhours{} hours across
\numscenes{} scene classes with three tiers of proxies. At the output end, the L1
attention tier (\hoursLone~h) infers gaze from headband camera video, exposing the
operator's next object of attention and
reach before the hand moves. Midway, the L2 exertion tier (\hoursLtwo~h) records
surface EMG from armbands, making load expectations readable before execution is
complete.
At the origin, the L3 cognition tier (\hoursLthree~h) records scalp EEG, giving
early readouts of task decisions and monitoring. Each tier inherits all proxies
from the tiers below. The dataset further provides multi-view RGB images, depth
maps and point clouds, 6-DoF head and wrist pose, and step-level action and
object annotations. The controlled L3 setting additionally measures gaze, hand
pose and finger kinematics directly. Clock and calibration data are released
with the dataset so that any read rate can be re-aligned. The three tiers cover
three stations, from visual orienting to cortical preparation,
making the process behind visual--action pairs available for modeling at dataset
scale. Project page: \url{https://flagopen.github.io/HumanCOIN}.
}
\reportbibstyle{unsrtnat}
\AtBeginDocument{\hypersetup{pdfauthor={\authormetadata}}}

\begin{document}
\newsavebox{\arxivcoverassets}
\sbox{\arxivcoverassets}{%
  \includegraphics{baai_logo.pdf}%
  \includegraphics{cover_overview.pdf}%
}
\reportfrontmatter
\section{Introduction}
\label{sec:intro}

Egocentric data has become the primary route from human video to embodied
learning. A head-mounted camera records the scene from the operator's viewpoint,
hand pose recovers the motion of the hands, and together they form a
visual--action trajectory: what was seen and how the hand moved. This paradigm
has scaled from passive everyday video~\cite{grauman2022ego4d,damen2020rescaling}
to instrumented manipulation corpora~\cite{sener2022assembly101,fan2023arctic,
hoque2026egodex} reaching tens of thousands of
hours~\cite{egosuite2026,xperience10m2026}, and the resulting data has been used
directly to train robot policies~\cite{kareer2024egomimic,yang2025egovla,
luo2025beingh0}. Across all variations in scale and form, the basic unit remains
the observation--action pair $(o, a)$.

\begin{wrapfigure}{r}{0.55\linewidth}
  \centering
  \includegraphics[width=\linewidth]{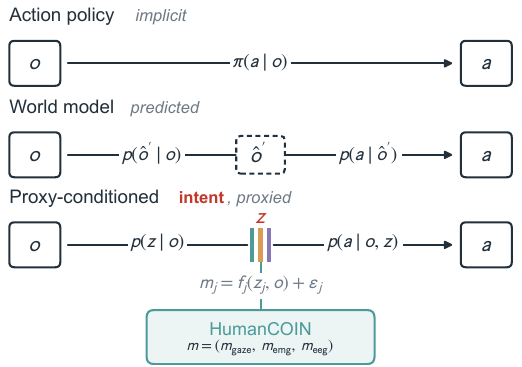}
  \caption{The middle of $o \to a$, and what each route puts there. An action
  policy models $\pi(a\mid o)$ directly, leaving the middle implicit in its
  weights; a world model inserts a predicted future $\hat{o}'$ and factors
  through it. The third row is not another method: \dataset{} supplies the
  middle as proxies $m=(m_\mathrm{gaze}, m_\mathrm{emg}, m_\mathrm{eeg})$
  for the intent $z$, on a common time base with the action.
  \label{fig:motivation}}
\end{wrapfigure}

Two broad approaches have organised work around this unit. The first directly
models $\pi(a \mid o)$: all available information---visual observations, depth,
pose, and language---is subsumed into $o$, and the mapping to $a$ is learned
from data at scale~\cite{brohan2023rt2,kim2024openvla,black2024pi0,oneill2024openx,
khazatsky2024droid,robocoin2025}; the operator's intent state is compressed
wholesale into network weights, rendering it unobservable---when a policy fails,
there is no way to localise whether the target was wrong, the force incorrect, or
an error signal missed. The second introduces an explicit intermediate and chooses
the future: world models and video--action co-models predict a latent
representation $\hat{o}'$ of the next observation, then generate the action,
$\pi\colon o \to \hat{o}' \to a$~\cite{du2023unipi,wu2024gr1,hu2025vpp,
li2025uva}; however, a predicted future records only the visible outcome of an
action, not the operator state that produced it, and therefore diverges from
intent precisely where supervision is needed most. \cref{fig:motivation} sets
the two routes against the alternative: both place something in the middle of
$o \to a$, and what they place there is either folded away or inferred, never
read off the operator. What is missing is not another estimate of that middle
but a record of it, and that is what \dataset{} provides.

Intent cannot be observed directly, but leaves measurable physical traces along
the chain from decision to effect---a chain we term the \emph{intent
pathway}---and we denote the operator's state along this pathway as the latent
variable $z$. The standard policy implicitly integrates over $z$,
$p(a \mid o)=\int p(a \mid o,z)\,p(z \mid o)\,\mathrm{d}z$, folding it into
weights and rendering it unobservable; recording proxies for $z$ would allow
both factors---$p(m \mid o)$, a proxy model, and $p(a \mid o,m)$, the
proxy-conditioned policy---to be trained independently.
Consider lifting a cup: gaze settles on the rim before the hand departs,
externalising intent as a visual orienting signal
($m_\text{gaze}$~\cite{land2001,hayhoe2005,johansson2001}); forearm muscles then
load grip force to the anticipated weight before the cup leaves the surface,
encoding a prior over object properties as a motor command
($m_\text{emg}$~\cite{johansson1988weight,johansson2009coding}); and cortical
activity has already been building a Bereitschaftspotential for hundreds of
milliseconds, with an error-related negativity ready to fire within milliseconds
of a mismatch ($m_\text{eeg}$~\cite{kornhuber1965,shibasaki2006bp,gehring1993ern}).
Each proxy precedes $a$ and has mature wearable sensors with decades of
signal-processing precedent~\cite{farina2014extraction,wolpaw2002brain};
synchronised recordings exist~\cite{lin2026egobrain}, yet all remain confined to
laboratory settings with few participants and short scripted trials. Large-scale
synchronised recording of all three proxies in naturalistic manipulation scenes
has not been achieved.

We present \dataset{}, a large-scale egocentric manipulation dataset that
records a proxy vector $m=(m_\text{gaze},m_\text{emg},m_\text{eeg})$ for $z$
alongside every visual--action pair, making both $p(m \mid o)$ and
$p(a \mid o, m)$ supervisable from data. The corpus
comprises 12,500 hours across \numscenes{} scene classes, collected at \numsites{} sites.
It is structured into
three tiers (\cref{fig:stats}a), one per proxy:

\begin{itemize}
  \item \textbf{L1, Attention tier (\hoursLone{}~h).} Proxy $m_\text{gaze}$:
  multiview egocentric RGB with inferred gaze, recording the output of visual
  orienting decisions.
  \item \textbf{L2, Exertion tier (\hoursLtwo{}~h).} Proxy $m_\text{emg}$:
  bilateral forearm surface EMG, recording the peripheral
  projection of motor commands; the signal envelope reflects muscle activity from
  which grip force and load may be inferred before object motion.
  \item \textbf{L3, Cognitive tier (\hoursLthree{}~h).} Proxy $m_\text{eeg}$:
  scalp EEG, recording movement preparation potentials and
  error-related potentials that precede observable motion.
\end{itemize}

The tiers are nested in instrumentation: each inherits every proxy of the tier
below and takes on one more, each proxy being a suite of instruments rather than a
single device. Depth is a design axis, not a separate collection: the families that
survive the same canonical filters at every tier recur at all three measurement
depths on a shared time base. Scene, objects, posture, and protocol nonetheless
move with depth, so a cross-tier comparison is task-matched rather than
condition-matched; \cref{sec:depth} states what the shared families hold fixed and
what they leave open. The three
proxies are recorded by the three tiers, and the action $a$ itself is captured by
the egocentric video every tier carries: $z$ remains latent, but $m$ is recorded
rather than marginalised. Beyond the
tier-specific modalities, the dataset provides depth maps and 3D point clouds
reconstructed from egocentric views, 6-DoF head and wrist pose with finger
kinematics, and step-level action--object annotations, all registered to a single
coordinate frame and time axis. L3 serves as a multimodal reference in its controlled
setting: there, $m_\text{gaze}$, head and wrist pose, and finger kinematics are
measured directly with dedicated instruments, while their counterparts at L1, at L2, and
in the open L3 subset are predicted from multiview video, aligning all three tiers in semantics, coordinate
frame, and time base so that the controlled measurements provide per-frame
reference targets for shallow-tier predictors.

Relative to existing corpora, \dataset{} differs not in scale but in coverage:
three proxy depths of the intent pathway under one task set and one time base,
making the generative process behind each visual--action pair available for
modelling rather than only for imitation. The dataset is released under a research licence (CC BY-NC 4.0) together with acquisition software, a data browser and downloader, and a
ready-to-use client library. Our contributions are threefold:

\begin{itemize}
  \item \textbf{A proxy-grounded acquisition paradigm.} Three wearable proxies
  of the intent pathway are recorded alongside first-person manipulation
  and aligned to a common time base, making measurement depth a design axis
  (\cref{sec:acquisition}).
  \item \textbf{A large-scale intent-pathway dataset.} The corpus comprises
  \numhours{} hours over \numscenes{} scene classes, organised into
  three tiers, each contributing one proxy of the pathway, sharing a common task ontology and time base
  (\cref{sec:distribution}).
  \item \textbf{An open ecosystem for the community.} Acquisition software, a
  data browser and downloader, and an out-of-the-box client library are released
  with the corpus (\cref{sec:storage}).
\end{itemize}

\section{Related Work}
\label{sec:related}

\subsection{Egocentric Vision Corpora}

Egocentric video has been scaled along two axes, hours and annotation density. Ego4D~\cite{grauman2022ego4d} collected 3,670 hours from 931 participants in nine countries and defined benchmarks for episodic memory, forecasting, and social interaction. EPIC-KITCHENS~\cite{damen2018epic} traded breadth for density, annotating cooking with verb--noun segments; its hundred-hour extension, EPIC-KITCHENS-100~\cite{damen2020rescaling}, added 89K action segments. EGTEA Gaze+~\cite{li2020eye} recorded eye movements over 28 hours of cooking and turned attention into a label. Ego-Exo4D~\cite{grauman2024egoexo4d} added a synchronized third-person view: 1,286 hours from 740 participants, annotated with skills, keysteps, and proficiency.

Assembly101~\cite{sener2022assembly101} instrumented tabletop assembly with 4,321 multi-view videos; ARCTIC~\cite{fan2023arctic} captured bimanual manipulation of articulated objects with 2.1M frames and 3D hand--object annotations; HOI4D~\cite{liu2022hoi4d} recorded 2.4M RGB-D frames of category-level hand--object interaction; and HoloAssist~\cite{sener2023holoassist} annotated 166 hours of human--assistant collaboration, including mistakes and their corrections.

A newer generation raises the bar on geometry and duration: Nymeria~\cite{ma2024nymeria} captures multimodal daily motion with full-body and hand tracking, HD-EPIC~\cite{perrett2025hdepic} adds 3D object and gaze grounding to kitchen annotation, EgoDex~\cite{hoque2026egodex} records 829 hours of manipulation with on-device 3D hand tracking, and EgoLife~\cite{yang2025egolife} follows six participants for a week. All remain external observations: none records muscle or cortex during the act.
EgoSuite-Open100K~\cite{egosuite2026} has opened the first 10,000 hours of a planned 100,000-hour programme of egocentric manipulation, with 3D hand and body pose over 10,000 tasks and 500 environments, and Xperience-10M~\cite{xperience10m2026} releases 10,000 hours and 2.88B RGB frames of in-the-wild human experience with depth, camera pose, and captions. Both record the outside of the action at a scale no research corpus reaches. Neither records gaze, muscle, or cortex, and neither pairs the same task with a shallower version of itself.

\subsection{Physiological and Neural Sensing}

Surface EMG has long served as the interface between motor intent and machine.
Farina et al.~\cite{farina2014extraction} review how motor-unit activity is
extracted from the surface signal and used to decode intended movement and force,
and Ninapro~\cite{atzori2014ninapro} standardised forearm recordings across
subjects. Grip-force studies show that forearm muscles load to the anticipated
object weight before the object leaves the surface~\cite{johansson1988weight,
johansson2009coding}, encoding the operator's prior over object properties as an
observable EMG signal before any visual feedback is available. EgoEMG~\cite{egoemg2026} pairs bilateral forearm EMG with egocentric video for hand pose estimation.

EEG-based interfaces have decoded motor preparation, error monitoring, and
cognitive load for decades~\cite{wolpaw2002brain}. Kornhuber and Deecke
identified the Bereitschaftspotential, a slow cortical build-up that precedes
voluntary movement by hundreds of milliseconds~\cite{kornhuber1965}, and
Shibasaki and Hallett provide a systematic account of its neural
basis~\cite{shibasaki2006bp}. Gehring et al.\ established the error-related
negativity (ERN): cortex produces a discriminable response within milliseconds of
an incorrect outcome, before any external feedback~\cite{gehring1993ern}. Taking
EEG outside the laboratory is complicated by motion artefacts; the mobile
brain/body imaging paradigm~\cite{gramann2011mobilebrainbody} and active-electrode
hardware have made wearable acquisition feasible, though signal quality remains
below that of seated recordings.

Gaze provides the earliest behavioural signature of intent: it reaches the target
of a pointing movement before the hand does~\cite{neggers2000ocular}, and
appearance-based trackers~\cite{krafka2016gazecapture} made the signal recordable
outside the laboratory; in everyday tasks fixations run ahead of the hand and
mark the objects about to be acted on~\cite{land2001,hayhoe2005}; and in object
manipulation gaze reaches the contact point before the fingers
close~\cite{johansson2001}. Pupil diameter adds a physiological readout of
processing load~\cite{beatty1982}. When gaze and EEG are recorded together,
fixation-related potentials lock cortical activity to the currently fixated
object~\cite{dimigen2011}, showing that the three signals are mutually
constraining rather than independent.

The neuroergonomics programme~\cite{parasuraman2007neuroergonomics} argued that
these measurements belong inside real tasks. Most studies nevertheless remain
laboratory experiments: the task is designed to elicit a signal, scenes are
controlled, participants few and trials short, so behaviour, muscle, and cortex
are rarely recorded on one timeline. EgoBrain~\cite{lin2026egobrain} pairs a
32-channel EEG headset with egocentric video over 61 hours from 40 participants,
the first corpus to put cortex on the same timeline as first-person video; it is
seated and scripted, however, covering 29 daily actions in one setting, with no
gaze, no muscle, and no scene geometry. Large-scale synchronised recording of all
three signals in naturalistic manipulation scenes has not been achieved.

\subsection{Human Data for Robot Learning}

Human video has long been used to teach robots, but the methods differ in what they inherit from the human.
TCN~\cite{sermanet2018tcn} learned a self-supervised video representation for imitation, DexMV~\cite{qin2022dexmv} brought human video and estimated hand pose to dexterous manipulation, R3M~\cite{nair2022r3m} pretrained a visual representation on Ego4D and transferred it to manipulation, VRB~\cite{bahl2023vrb} extracted affordances from human videos as a representation for robot policies, and EgoMimic~\cite{kareer2024egomimic} scaled imitation learning from egocentric human video captured with a portable rig. These methods inherit appearance and affordance from the human, but not the signals that produced the action.

A second line introduces an explicit intermediate between $o$ and $a$ by
choosing the future. UniPi~\cite{du2023unipi} casts the policy as a video
diffusion model and reads actions from the generated frames; GR-1~\cite{
wu2024gr1} jointly predicts future frames and actions in a transformer; VPP~\cite{
hu2025vpp} and UVA~\cite{li2025uva} condition the policy on latent
representations of predicted future observations. These models make the gap
between $o$ and $a$ explicit, but the intermediate they supervise is a predicted
future observation rather than the operator's current state: the two diverge when
an attempt fails or is abandoned, and neither grip force nor cortical preparation
is recoverable from a predicted frame.

A third line adds modalities to the policy itself. Any2Policy~\cite{any2policy2024}
trains a modality-agnostic encoder that accepts text, audio, images, or point
clouds in any combination and connects them to a policy head.
Chain-of-Modality~\cite{chainofmodality2025} prompts a vision-language model to
chain modalities from human demonstration videos---images, armband EMG, and
audio---into task plans with explicit control parameters. Play to the
Score~\cite{feng2024play} fuses multiple sensory streams with stage-guided
dynamic weighting during manipulation. These works show that extra modalities
help a policy, and some already incorporate EMG at the policy end; none, however,
is grounded in a corpus that systematically records the operator's intent proxies
across multiple depths under a shared task set and time base.

\section{Data Distribution}
\label{sec:distribution}

\begin{figure}[!t]
  \centering
  \includegraphics[width=\linewidth]{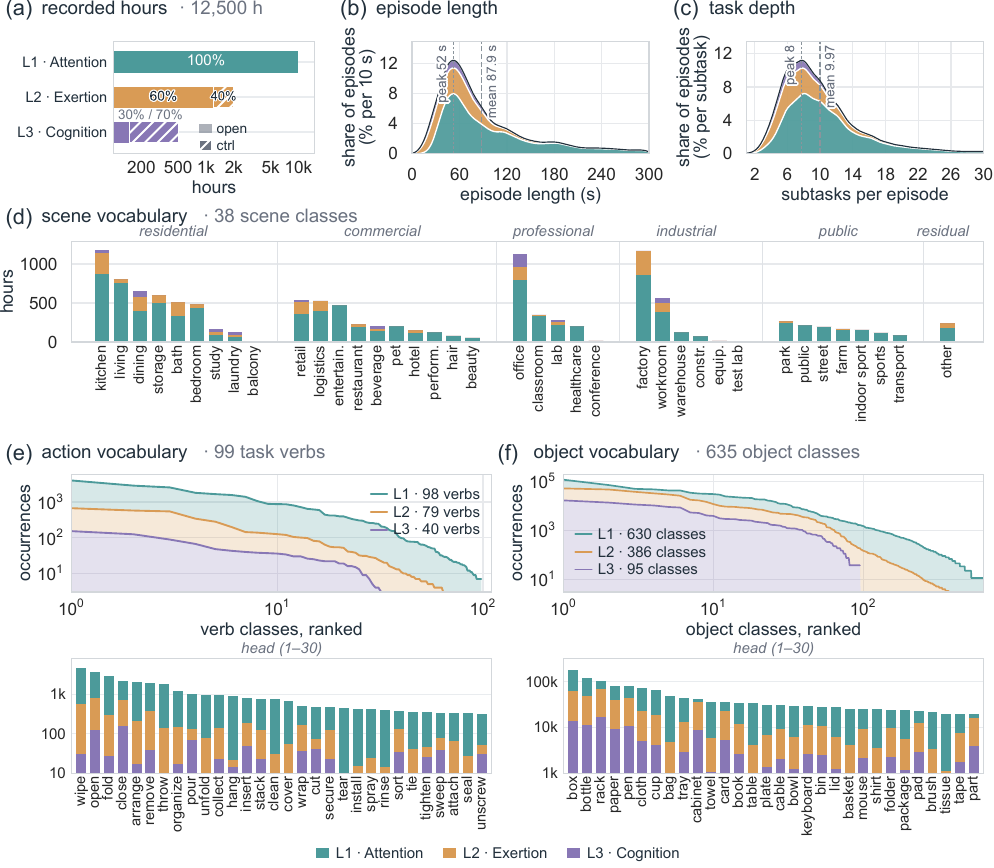}
  \caption{HumanCOIN in one view. (a)~The three tiers and how their hours split: \numhours\,h in total, at \hoursLone{} / \hoursLtwo{} / \hoursLthree\,h (\shareLone{} / \shareLtwo{} / \shareLthree\%) for L1 / L2 / L3, drawn on a log axis; solid segments are unscripted (open) recording, hatched segments scripted or bench-controlled. (b)~Episode length and (c)~task depth: stacked kernel densities of episode duration and of subtasks per episode, with the peak and the mean marked. (d)~All \numscenes{} scene classes, grouped into the \numzones{} zones and a residual class, with the tier split of each class's hours. (e)~The task verbs: the upper panel ranks each tier's occurrences against vocabulary rank, and the lower panel gives the thirty most frequent verbs on a log occurrence axis. (f)~The \numobjects{} object classes, ranked by frequency, with the thirty most frequent below, again on a log axis. Every stacked bar is split by tier.}
  \label{fig:stats}
\end{figure}

\dataset{} holds \numhours{} hours of egocentric manipulation data, organised
along three measurement depths: each tier corresponds to one observable proxy
$m_\text{gaze}$, $m_\text{emg}$, or $m_\text{eeg}$, from the visual orienting
signal at the pathway's distal end, through the peripheral motor command, to
cortical preparatory activity. The corpus spans \numscenes{} scene classes,
\numsubverbs{} task verbs, and \numobjects{} object classes. The three tiers share
a common task ontology and time base; scene and behavioural diversity unfold within
each tier.

\subsection{The Depth Axis}
\label{sec:depth}

Existing egocentric corpora record the two ends of an operation: the observation
$o$ gives the visual environment the operator inhabited, and the action $a$ gives
the motion and effect of the hands---together, what happened. These two ends form
the baseline modality set $\mathcal{M}_0 = \{o, a\}$ that all existing corpora
share. What $\mathcal{M}_0$ does not answer is why a particular action occurred
in a particular way at a particular moment.

We use \emph{intent} as a working term for the operator's task-relevant state,
denoted $z$. $z$ is neither a synonym for $o$ nor an early version of $a$; it
lies between perception and execution and shapes how an action unfolds. For
any $o,a$,
\begin{equation}
  p(a \mid o) = \int p(a \mid o, z)\, p(z \mid o)\, \mathrm{d}z.
  \label{eq:intent}
\end{equation}
\cref{eq:intent} writes $p(a \mid o)$ as an integral over $z$: given $o$,
the distribution of $a$ is determined jointly by the posterior $p(z \mid o)$ and
the action model $p(a \mid o, z)$. It supplies no directly observable quantity:
$z$ itself is not observed. The question is not whether $z$ exists, but whether
$z$ carries information about $a$ beyond $o$, i.e., whether $p(a \mid o, z)$
differs from $p(a \mid o)$. We do not assume $a \perp o \mid z$; the observation
remains in the policy.

Because $z$ is latent, \cref{eq:intent} cannot be used directly for training
or evaluation. The corpus records proxies for $z$ at three stages along the intent pathway,
$m=(m_\text{gaze},m_\text{emg},m_\text{eeg})$, not $z$ itself. A proxy is a lossy
measurement of $z$:
\begin{equation}
  m_j = f_j(z_j, o) + \varepsilon_j, \qquad
  j \in \{\text{gaze}, \text{emg}, \text{eeg}\}.
  \label{eq:proxy}
\end{equation}
$f_j$ reads out only the part of $z_j$ that the sensor can reach, and
$\varepsilon_j$ is noise; $f_j$ is not required to be invertible, nor $m_j$ to
contain all of $z_j$. The corpus therefore supports a proxy-conditioned policy
$p(a \mid o, m)$, not $p(a \mid o, z)$. This distinction runs through the paper:
$z$ is the explanatory frame, $m$ is what the data contain.

$m$ is not a sufficient statistic for $z$. The three proxies differ in noise,
latency, and coverage; they are not interchangeable, nor are they nested
estimates of a single scalar. Together they cover three observable stations,
from visual orienting to cortical preparation and monitoring, but they do not
exhaust $z$. We do not claim that these proxies identify $z$; the limitation is
discussed in \cref{sec:discussion}.

\dataset{} asks one measurement question at each station:

\begin{tcolorbox}[enhanced, colback=thmbg, colframe=thmframe, boxrule=0.5pt,
  arc=2pt, left=4pt, right=4pt, top=4pt, bottom=4pt]
{\small
\textbf{Q1.}\; Where does gaze land before the hand departs, and how early does it carry information about the upcoming action?\par\vspace{3pt}
\textbf{Q2.}\; How does the effector respond to the motor command before the action occurs, and what does the EMG envelope carry about grasp type and load?\par\vspace{3pt}
\textbf{Q3.}\; What cortical activity precedes the motor command, and what does it carry about preparation and error monitoring?}
\end{tcolorbox}

The three tiers are not independent datasets but a ladder nested in
instrumentation: each tier inherits every proxy of the tier below and takes on one
more---gaze, EMG, EEG---and every proxy is a suite
of instruments, whose members the controlled setting can upgrade from estimation to
measurement. Their hours---\hoursLone, \hoursLtwo, and \hoursLthree---decrease
with depth, reflecting the higher acquisition cost of deeper proxies.

\begin{itemize}
\item\textbf{L1, Attention tier (\hoursLone~h).} Fourteen heterogeneous
headbands from thirteen vendors record multiview egocentric RGB; gaze is estimated from video and depth is reconstructed from the front
stereo pair. L1 is collected in open scenes, where operators move freely and
tasks are not scripted. The tier has neither EMG nor EEG.

\item\textbf{L2, Exertion tier (\hoursLtwo~h).} The headband is fixed to a single
model, and bilateral forearm surface EMG is added. Gaze is still estimated from video; scenes and objects become
scripted, and tasks follow a task sheet. The tier has no EEG.

\item\textbf{L3, Cognition tier (\hoursLthree~h).} The cortical proxy is added,
together with the L3 kit: two wrist views and a third-person view, and, in the
controlled setting, the eye tracker, the trackers, and the gloves that measure
gaze, hand pose, and finger kinematics directly. L3 has two settings: \emph{open}
uses a lightweight EEG configuration in natural scenes; \emph{controlled} uses a
high-density cap at a fixed desk, with imagery before execution. By the split used
in \cref{fig:stats}a, about \num{350}\,h of L3 are controlled and \num{150}\,h
open; only the controlled part supplies direct measurements.
\end{itemize}

Thus the gaze, hand pose, and finger kinematics of L1, L2, and open L3 are
estimates; the same quantities in the controlled L3 subset are measurements. That
subset therefore provides the reference measurements $y$ against which the
estimates $m$ can be compared frame by frame on the same episode. ``L3 as reference'' means
that $y$ is a reference measurement for $m$, not ground truth for $z$.

The tiers are nested in instrumentation but not in distribution: scene, objects,
posture, and protocol move from open to controlled with depth. A cross-tier
comparison is therefore task-matched, not condition-matched. What the tiers share
is a core of task families: those that survive both canonical filters,
$\mathcal{C}_1$ and $\mathcal{C}_2$, and across them the task is the same at every
depth while the staging is not---L1 instances are unguided, and the object set,
step order, posture, and protocol are fixed only from L2 onward. The probe sets
$\mathcal{E}_2$ and $\mathcal{E}_3$ are the increments and need not be shared. What they do not share
is a condition: the corpus does not pair a family with one scene, posture, and
protocol across tiers. A depth-only comparison---same scene, objects, posture, and
task, only the instruments changed---is thus a construction the corpus admits, not
a split it releases: one fixes a family and its staging, takes the episodes each
tier recorded for it, and reports the residual differences rather than assuming
them away.

Beyond modalities, the tiers follow an open-to-controlled gradient in task,
scene, and acquisition paradigm: the deeper the tier, the more controlled and the
narrower its coverage. \emph{Open} means the task happens in a natural scene in
the operator's own way---objects occur as they occur, posture is unconstrained,
recording runs free. \emph{Controlled} means the scene, object set, posture, and
acquisition protocol are fixed one by one, trading behavioural freedom for
reproducibility. The design keeps one task comparable across depths while the
corpus spans natural behaviour to controlled conditions.

In task, a \emph{family}---a goal, an object set, and a dominant action sequence,
coarse enough to recur at every tier---enters a tier only after canonical
filtering, and each tier adds a probe set targeting the proxy it introduces:
\begin{equation}
\begin{aligned}
\mathcal{T}_1 = \mathcal{U},\quad
\mathcal{T}_2 = \mathcal{C}_1(\mathcal{T}_1)\cup\mathcal{E}_2,\quad
\mathcal{T}_3 = \mathcal{C}_2(\mathcal{T}_2)\cup\mathcal{E}_3,
\end{aligned}
\label{eq:tiersets}
\end{equation}
where $\mathcal{T}_k$ is the family set at tier $k$, $\mathcal{U}$ the natural
repertoire, and $\mathcal{C}_1$, $\mathcal{C}_2$ the two canonical filters:
$\mathcal{C}_1$ fixes the object set, placement, and step order; $\mathcal{C}_2$
additionally fixes posture and phases (imagery before execution in the controlled
setting). The probe sets are the tier's increment: muscle-load tasks at
L2---heavy objects, balancing, and tactile discrimination
($\mathcal{E}_2$)---place the effector at a loaded working point; high-cognition
tasks at L3---instruction-following, long-horizon stacking, and error-prone
insertion ($\mathcal{E}_3$)---impose complex multi-stage demands on the cortex.

\begin{wraptable}{r}{0.55\linewidth}
  \centering
  \caption{The four axes that separate the three tiers. Each tier inherits every modality class of the tier below; the four axes co-vary with depth, so the tiers form a design gradient rather than a controlled experiment.}
  \label{tab:tiers}
  \small
  \setlength{\tabcolsep}{3pt}%
  \begin{tabular}{@{}p{0.17\linewidth} >{\centering\arraybackslash}p{0.15\linewidth} >{\centering\arraybackslash}p{0.23\linewidth} >{\centering\arraybackslash}p{0.35\linewidth}@{}}
    \toprule
     & L1 & L2 & L3 \\
    \midrule
    Scene & open & staged & staged + desk \\
    Objects & natural & specified & specified \\
    Posture & free & free & still \\
    Protocol & free & phased & phased + imagery \\
    \bottomrule
  \end{tabular}
\end{wraptable}

Four dimensions move together from open to controlled: scene from fully open to
staged to a fixed desk; objects from natural occurrence to a specified set with
each item measured for colour, shape, size, material, and weight, so that object weight
becomes a known quantity; posture from unconstrained to as still as possible,
since electrode contact and eye tracking degrade with movement; and acquisition
from free recording to phased protocols with an imagery stage, so that motor
preparation is not buried by the movement that follows.
\cref{tab:tiers} sets the four dimensions side by side. These four axes
co-vary with depth, so they describe a design gradient rather than a controlled
experiment; isolating depth requires the depth-only comparison described above.

\begin{tcolorbox}[enhanced, colback=thmbg, colframe=thmframe, boxrule=0.5pt,
  arc=2pt, left=4pt, right=4pt, top=4pt, bottom=4pt]
{\small
\textbf{Scope.}\; \dataset{} records proxies
$m=(m_\text{gaze},m_\text{emg},m_\text{eeg})$ for the latent state $z$, not
$z$ itself. It provides time-aligned multimodal streams, L3 controlled reference
measurements $y$, and per-episode calibration and alignment.}
\end{tcolorbox}

\subsection{Episode Duration and Task Stages}

Episode duration determines how much context and how many steps a single sample carries, and it largely constrains the kind of training and evaluation the sample can support. The duration of \dataset{} is not set uniformly; it varies systematically with tier, task type, and difficulty.

\noindent\textbf{Duration distribution (\cref{fig:stats}b).} The distribution is unimodal with a long tail: the density peaks at \secPeak\,s and the mean sits at \secMean\,s, with the shortest episodes near \secShort\,s and an open tail beyond \secTail\,min. The tiers differ systematically. L1 has the widest range and the heaviest tail, at a mean of \secMeanLone\,s; L2 and L3 are more concentrated, at \secMeanLtwo{} and \secMeanLthree\,s. That gap belongs to the recording protocol rather than to the tasks: L1 is segmented after the fact from long continuous recordings, whereas L2 and L3 are scripted episodes with bounded sessions; electrode stability further shortens the deepest sessions. The gap is therefore not evidence about task difficulty, nor about depth as such.

\noindent\textbf{Task stages (\cref{fig:stats}c).} Subtasks per episode count the stages a task is composed of rather than the temporal granularity of annotation. The count is highest at L1, at \subMeanLone{} per episode, where natural activity is split into many short steps; it falls to \subMeanLtwo{} at L2 and \subMeanLthree{} at L3, where a fixed protocol groups the work into fewer, longer stages.

Overall, \dataset{} shows clear tier differences along both duration and task structure: the shallow tiers hold long, loosely segmented, and diverse everyday activity, while the deep tiers concentrate on scripted, standardized tasks in which the same work is organized into fewer, longer stages. Duration characterizes the recording window, stage count characterizes task structure, and together they form the corpus's spectrum from natural behaviour to complex controlled tasks.

\subsection{Scene Distribution}

The value of egocentric data depends largely on how far it generalizes in the real world, so the scene distribution should not be trimmed for convenience of collection. The \dataset{} scene vocabulary holds \numscenes{} classes, grouped into \numzones{} zones---residential (\zoneResidential), commercial (\zoneCommercial), professional (\zoneProfessional), industrial (\zoneIndustrial), and public (\zonePublic)---plus \zoneResidual{} residual class. \cref{fig:stats}d arranges them by zone and by tier, with the hours of each class stacked to show how the three tiers split the total; \appref{app:scenes} lists all \numscenes{} with their tier coverage.

\noindent\textbf{A residential centre of gravity.} Residential holds \hoursResidential{} hours; commercial, \hoursCommercial; professional, \hoursProfessional; industrial, \hoursIndustrial; public, \hoursPublic; and the residual, \hoursResidual. The ordering is not incidental: manipulation is densest where people live and work, and the corpus follows that density rather than flattening it. Within each zone a few classes carry most of the hours---residential runs from the kitchen's \hoursKitchen{} hours down to the balcony's \hoursBalcony, while the industrial zone is led by the factory workshop at \hoursFactory{} and the professional zone by the office at \hoursOffice. Across the whole corpus the five largest classes take 39\% of the hours, a concentration that mirrors how often those spaces are actually used.

\noindent\textbf{Selective coverage by tier.} L1 holds all \numscenes{} classes, L2 holds \numscenesLtwo{}, and L3 holds \numscenesLthree{}, with the remaining \numscenesAbsent{} absent from L3; the \numscenesLthree{} L3 classes account for \pctScenesLthreeOfLone\% of L1 hours and \pctScenesLthreeOfLtwo\% of L2. The scene set therefore contracts at each step---38 $\to$ 21 $\to$ 9, or 100\% $\to$ \pctScenesLtwo{} $\to$ \pctScenesLthree\% of the vocabulary---which is why the deep tier reads as a deliberate instrumented subsample rather than a miniature of the whole. Its nine classes are weighted toward the office, the dining room, the workroom, and the kitchen---spaces that admit a fixed desk, stable lighting, and a bounded session---while the controlled subset is narrower still, concentrated on the desk.

This is not a sampling bias but a direct consequence of the controlled setting: a high-density cap, an eye tracker, and a fixed desk require a fixed station, stable lighting, and a bounded session, and scenes such as the street, the shop floor, and the farm cannot physically accommodate them. The nine L3 classes are the classes with any L3 hours; the controlled subset is narrower still, concentrated on the desk. L1 is collected in open scenes, L2 in staged ones, and L3 spans both settings. The three tiers therefore differ in what they can record, and the scene distribution is a direct reflection of that difference.

Taken together, the scene distribution expresses a trade-off between diversity and record quality. From L1 to L3, coverage contracts from all \numscenes{} indoor and outdoor classes to a concentration on the \numscenesLthree{} classes that can be instrumented. The result is a corpus that spans from natural behaviour to complex controlled tasks, with task families that recur across tiers observed at several measurement depths.

\begin{figure}[t]
  \centering
  \includegraphics[width=\linewidth]{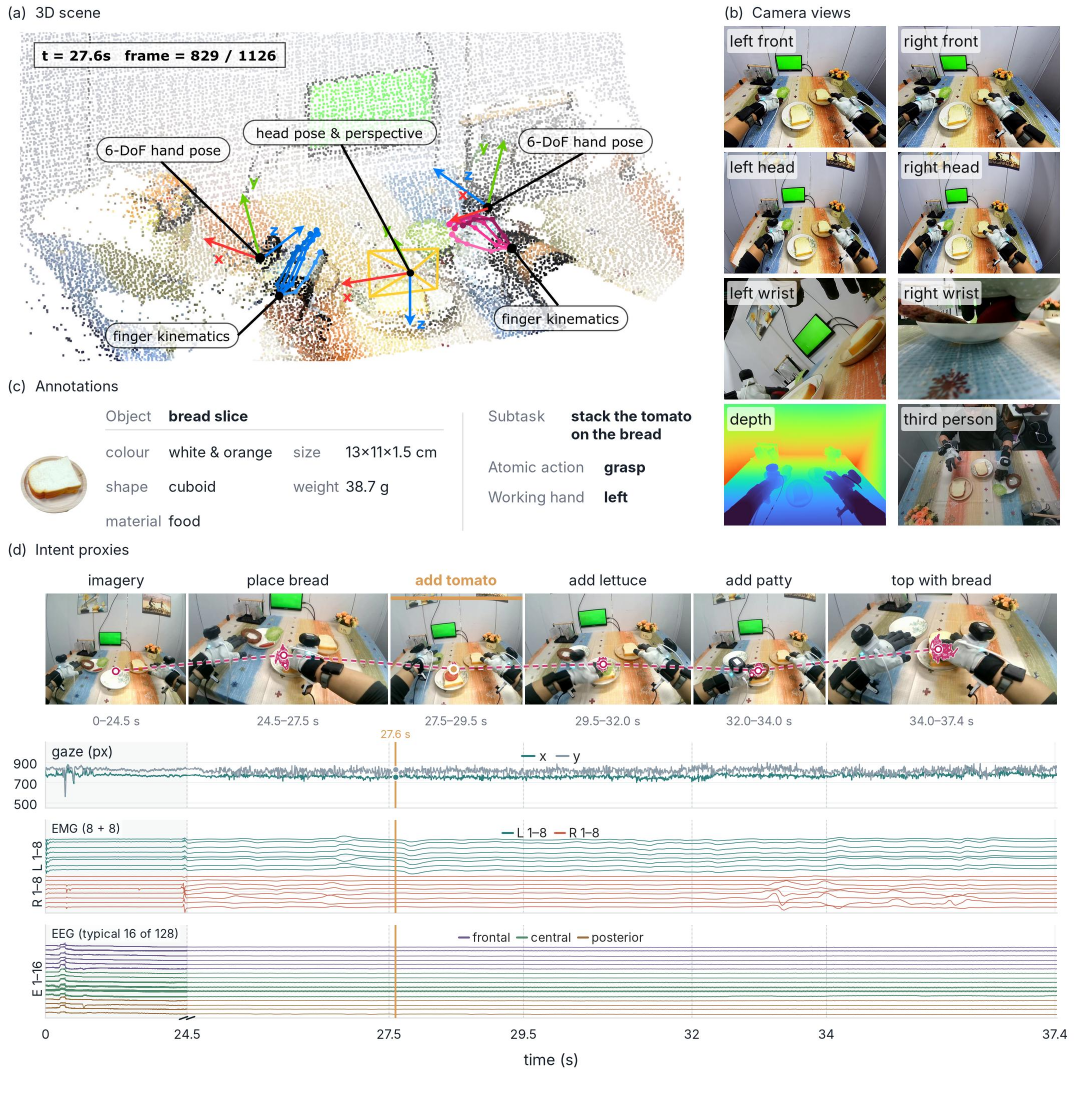}
  \vspace{-6ex}
  \caption{One controlled L3 episode in full, at $t=27.6$\,s (frame 829/1126). (a)~The metric point cloud, coloured by the head-camera RGB, with the head pose, both hand skeletons and their 6-DoF poses overlaid in one shared frame. (b)~Eight synchronized camera views arranged two per row: the four head-mounted views, the two wrist views, the metric depth map, and the third-person view. (c)~The annotation of the selected instance: the step sentence \emph{stack the tomato on the bread}, the measured object attributes (name, colour, shape, material, size, weight), the atomic action \emph{grasp}, and the working hand \emph{left}. (d)~The task stages and the three intent proxies of the same interval: gaze $(x,y)$, eight-plus-eight-channel EMG, and sixteen of the 128 EEG channels, on a shared time axis with the current instant marked; the eye reaches the object before the hand and the muscle load rises before contact, so each trace leads the action it announces. The stages are imagery, place bread, add tomato, add lettuce, add patty, and top with bread.}
  \label{fig:anatomy}
\end{figure}

\subsection{Task Distribution}
\label{sec:tasks}

A task entry is the unit of reuse: one captured segment, its ordered step sentences, and the objects it touches. The entries span cooking, assembly, cleaning, errands, and teaching, and differ in step count, object set, and manipulation difficulty; that variation is what allows the corpus to be split by domain, by horizon, or by difficulty instead of being used as one monolithic training set. Each step carries a task verb, the hand that performed it, and the objects it touches, so an entry can be read both as a plan and as a record of execution (\cref{fig:stats}e,f). The upper tiers are, with few exceptions, subsets of the lower tiers: scenes are strictly nested (L3 $\subset$ L2 $\subset$ L1); objects and verbs are nested except for a small probe increment---five objects first added at L2 and six verbs first added at L3. \cref{fig:stats}d--f shows the resulting vocabulary split.

\noindent\textbf{Action: \numsubverbs{} task verbs.} A task verb names what a step accomplishes as a goal---\emph{wipe}, \emph{open}, \emph{fold}---and is the unit a plan is written in. The vocabulary is open-ended in principle but compact in practice: \numsubverbs{} classes cover the step record, a median of \medSubverbClass{} occurrences per class, and the ten most frequent take \pctSubverbsTop\% of it. The head is everyday manipulation---wipe, open, fold, close, arrange---above a long tail of rarer verbs, so the same set serves both as a small label vocabulary and as a description fine-grained enough to distinguish what is being done. Behind each step sits a second, finer record: the \emph{motion verbs} that carry it out---\emph{pick}, \emph{place}, \emph{move}, \emph{grasp}---\nummotionverbs{} classes over the motion record, 2.5 per step.

The tiers treat the vocabulary differently. L3 keeps \numsubverbsLthree{} task verbs (40\%), and those verbs cover \pctSubverbsLthreeOfLone\% of L1 task occurrences. A protocol fixes the task set, so the verb list contracts to the tasks it scripts, while the steps that are recorded stay labelled in full: the deep tier is selective in vocabulary but not in the record.

\noindent\textbf{Object: \numobjects{} classes.} The object side is longer-tailed still: the ten most frequent classes take about \pctObjectsTop\% of the object occurrences, while the median class occurs only about \medObjectClass{} times in total, and the 46 most frequent classes already carry 80\% of the record (\cref{fig:stats}f). Coverage again separates the tiers: L1 holds \rawObjectsLone{} classes and L2 \rawObjectsLtwo{}, while L3 keeps \rawObjectsLthree{} and leaves the other \numobjectsAbsent{} absent from L3; those \rawObjectsLthree{} still account for \pctObjectsLthreeOfLone\% of L1 occurrences and \pctObjectsLthreeOfLtwo\% of L2. L3 is a subset of L2; L2 adds a small probe set of objects absent from L1. The deep tier thus keeps \pctObjectsKept\% of the vocabulary and more than four fifths of the record---the same contraction seen in the scenes, and for the same reason. \appref{app:objects} describes the measured object library---its size, weight, material and colour distributions---and how it is counted against this annotation vocabulary.

Taken together, all three vocabularies are long-tailed, and what the tiers trade is breadth against depth rather than vocabulary size: the shallow tiers spread the full vocabulary over everyday activity, while the deep tier contracts to the classes its protocol needs and concentrates the record there.

\subsection{Inside an Episode}

\cref{fig:anatomy} opens one L3 episode in full: the metric scene in \cref{fig:anatomy}a, the camera views and depth map in \cref{fig:anatomy}b, the annotation of the selected step in \cref{fig:anatomy}c, and the task stages with the intent proxies of the same interval in \cref{fig:anatomy}d. It is shown in the controlled L3 setting because that is where the body is instrumented: the gaze, the hands, and the wrists that the other tiers infer from video are measured here.

\noindent\textbf{Geometry.} The point cloud is reconstructed from the two front views; the head, both wrists, and the finger joints are placed in one metric frame, so a grasp is located on the object rather than only in the image (\cref{fig:anatomy}a). In the controlled L3 subset these quantities are measured: three trackers give the head and both wrists, and the gloves give the joints. At L1, at L2, and in the open L3 subset the same quantities are estimated from RGB, with each wrist root anchored in the scene frame through the depth map and the finger joints following in three dimensions. The frame is shared; the evidence behind it is not.

\noindent\textbf{Views.} Four egocentric views pair the two front lenses (left-front and right-front) with the left and right views, and at L3 two wrist views add the hands at work (\cref{fig:anatomy}b). A depth map is reconstructed on demand from the front pair, and in the controlled setting a third-person view records the operator and the desk from outside. The frames are stored as recorded, with no annotation drawn over the pixels.

\noindent\textbf{Annotation.} The selected instance is shown in \cref{fig:anatomy}c: each step carries a short description, a task verb, the hand that executed it, and the objects it touched, and at L2 and L3 the object record adds a photo, a hex colour, a shape, a material, a weight, and a size given in the parameters its shape calls for (radius and height for a cylinder, for instance). For the step in the panel the atomic action is \emph{grasp} and the working hand \emph{left}.

\noindent\textbf{Intent proxies.} Each episode is cut into stages, shown as the strip at the top of \cref{fig:anatomy}d. The three traces beneath the strip are the proxies for the state $z$ that the action is about to express, and each is read as a lead or a lag against the action it announces. Gaze is the eye's $(x,y)$ position, measured by the eye tracker in the pixels of its scene camera in the controlled L3 subset and estimated from the rectified head view at L1, at L2, and in the open L3 subset: the eye reaches the object before the hand does, so the trace externalises the target of the next reach---a fixation is a flat stretch, a saccade a near-vertical step, and the moment the eye settles marks what the hand is about to act on ($m_\text{gaze}$). EMG is eight envelope traces per forearm, flat at rest and rising before contact: the forearm loads grip force to the weight the hand expects, so a burst records the force the operator intends to apply, not only the force applied when the fingers close ($m_\text{emg}$). EEG is a stack of channel traces, three orders of magnitude below EMG and drawn on its own scale: the slow build before the movement and the sharp response after an error are the cortical preparation and the outcome check, read as deviations from the ongoing rhythm rather than as spikes ($m_\text{eeg}$). On one clock, the three answer what the instant means---where the operator is about to reach, how hard they expect to grip, and what was decided before the hand moved.

An episode is therefore one record aligned to a common time base and in a single frame: the views and the point cloud give the scene, the poses, joint angles, and gaze give the body's place in it, EMG and EEG give the exertion and the cortical preparation behind the movement, and the annotation gives each step its place in the task.

\subsection{Comparison with Existing Datasets}
\label{sec:compare}

\cref{tab:corpora} compares a representative set of these corpora on what they
record rather than on scale: it is a comparison, not a census, and the corpora
that share an axis with a row here are discussed above rather than tabulated. All of them observe the task from the head, and neither a third-person
view nor a wrist view (Ego-Exo4D, Nymeria, EgoSuite) adds more than the visible
outcome of manipulation. Only EgoBrain carries a neural signal, and it carries just
one (EEG); gaze is the only human signal shared by several corpora---Ego4D on a
subset, Ego-Exo4D in 99\% of takes, EGTEA and Nymeria throughout---and always in
isolation, so that where the
eye went and how hard the hand pulled have never been observed at the same moment.
\dataset{} records the core families at three instrumentation depths, which is what lets
``measured in the controlled L3 setting, predicted from video elsewhere'' hold for those families, and
a shallow-tier estimator be scored against the dataset's own reference measurements
rather than an external benchmark.

EgoBrain is a complement rather than a replacement. It has already shown that
cortex can be recorded on the first-person timeline, but does so in a controlled,
seated, scripted setting, trading breadth for signal density: 61 hours, 40
participants, 32 channels, 29 daily actions. \dataset{} takes the same premise
into naturalistic scenes and trades action variety for volume, holding 500 hours
at L3. What it adds is two things EgoBrain does not attempt: gaze and surface EMG
on a common time base, so that three proxies can be compared at one instant, and
scene geometry with hand kinematics, so that an action lands on an object rather
than staying inside the image.

Subtask annotation is already common under other names, but object annotation is
not: elsewhere an object is a name, while at L2 and L3 every instance is measured
for shape, material, size, and weight. That is a difference in capability, not in
labelling style---it lets an EMG envelope be read against a measured object weight rather
than an estimated one, so that the anticipation of grip force before contact can
be inferred from muscle activity rather than only assumed; grip force itself is
not recorded.

\begin{table}[!t]
  \centering
  \caption{Representative prior egocentric datasets against \dataset{}. $\checkmark$ measured, P predicted from video, R reconstructed from video, $\checkmark^{\ast}$ subset only, $\checkmark$/P measured where instrumented and predicted elsewhere, a cross not provided; 3D is a reconstructed depth or point cloud, Fingers joint-level hand tracking. Cameras counts per recording: head a head-mounted first-person camera, wrist a wrist-mounted one, exo a third-person camera. Frames at 30\,fps where only a duration is reported, and the \dataset{} count is master frames, one per 30\,Hz tick. $^{\dagger}$First 10{,}000\,h open; 2--6 head means the number of head-camera lenses varies between batches, always with a front stereo pair. Shaded rows are \dataset{} and its three tiers.}
  \label{tab:corpora}
  \small
  \setlength{\tabcolsep}{2pt}
  \renewcommand{\arraystretch}{1.18}
  \begin{tabularx}{\linewidth}{@{}l
      rc>{\raggedright\arraybackslash}p{1.5in}
      ccccc
      >{\centering\arraybackslash}X>{\centering\arraybackslash}X@{}}
    \hline
    Dataset & Hours & Frames & Cameras & 3D & Fingers & Gaze & EMG & EEG & Subtask anno. & Object anno. \\
    \hline
    Ego4D~\cite{grauman2022ego4d}               & 3{,}670 & $\approx$396M & 1 head (+224\,h multi-cam) & $\checkmark^{\ast}$ & \ding{55} & $\checkmark^{\ast}$ & \ding{55} & \ding{55} & $\checkmark^{\ast}$ & \ding{55} \\
    Ego-Exo4D~\cite{grauman2024egoexo4d}        & 1{,}286 & $\approx$139M & 1 head, 4--5 exo & R & $\checkmark^{\ast}$ & $\checkmark$ & \ding{55} & \ding{55} & $\checkmark$ & \ding{55} \\
    EPIC-KITCHENS~\cite{damen2020rescaling} & 100 & 20M & 1 head & \ding{55} & \ding{55} & \ding{55} & \ding{55} & \ding{55} & $\checkmark$ & \ding{55} \\
    EGTEA Gaze+~\cite{li2020eye}                & 28 & 2.5M & 1 head & \ding{55} & \ding{55} & $\checkmark$ & \ding{55} & \ding{55} & $\checkmark$ & \ding{55} \\
    Nymeria~\cite{ma2024nymeria}                & 300 & 201.2M & 1 head, 2 wrist, exo & R & \ding{55} & $\checkmark$ & \ding{55} & \ding{55} & $\checkmark$ & \ding{55} \\
    EgoDex~\cite{hoque2026egodex}               & 829 & 90M & 1 head & \ding{55} & $\checkmark$ & \ding{55} & \ding{55} & \ding{55} & \ding{55} & \ding{55} \\
    EgoSuite-Open100K~\cite{egosuite2026}       & 10{,}000$^{\dagger}$ & 1.08B & 1 head, 1 wrist$^{\ast}$ & R & $\checkmark$ & \ding{55} & \ding{55} & \ding{55} & $\checkmark$ & \ding{55} \\
    Xperience-10M~\cite{xperience10m2026}       & 10{,}000 & 2.88B & 6 head & R & $\checkmark$ & \ding{55} & \ding{55} & \ding{55} & $\checkmark$ & \ding{55} \\
    EgoBrain~\cite{lin2026egobrain}             & 61 & 6.6M & 1 head & \ding{55} & \ding{55} & \ding{55} & \ding{55} & $\checkmark$ & $\checkmark$ & \ding{55} \\
    \hline
    \textbf{\dataset{}}
      & \numhours & \annotTotal & 2--6 head, 2 wrist*, exo* & R & $\checkmark$ / P
      & $\checkmark$ / P & $\checkmark$* & $\checkmark$* & $\checkmark$ & $\checkmark$ \\
    \rowcolor{lgreen}\pc{L1\ \textperiodcentered\ Attention}
      & \pc{\hoursLone} & \pc{\annotLone} & \pc{2--6 head} & \pc{R} & \pc{P}
      & \pc{P} & \pc{\ding{55}} & \pc{\ding{55}} & \pc{$\checkmark$} & \pc{\ding{55}} \\
    \rowcolor{lgreen}\pc{L2\ \textperiodcentered\ Exertion}
      & \pc{\hoursLtwo} & \pc{\annotLtwo} & \pc{4 head} & \pc{R} & \pc{P}
      & \pc{P} & \pc{$\checkmark$} & \pc{\ding{55}} & \pc{$\checkmark$} & \pc{$\checkmark$} \\
    \rowcolor{lgreen}\pc{L3\ \textperiodcentered\ Cognition}
      & \pc{\hoursLthree} & \pc{\annotLthree} & \pc{4 head, 2 wrist, exo*} & \pc{R} & \pc{$\checkmark^{\ast}$}
      & \pc{$\checkmark^{\ast}$} & \pc{$\checkmark$} & \pc{$\checkmark$} & \pc{$\checkmark$} & \pc{$\checkmark$} \\
    \hline
  \end{tabularx}
\end{table}

\section{Hardware Configuration}
\label{sec:acquisition}

The intent proxies summarized above do not appear in a recording by themselves.
Each becomes a record only through an instrument that converts a physical
quantity into a measurable signal, so the kit is a set of \emph{measurement
proxies}: one class of instrument per
component of $z$, each chosen for what it can physically reach. Every instrument
keeps its own clock, its own native sampling rate, and its own coordinate frame.
This section describes what is worn at each tier, the two settings in which the
kit is deployed, the two-phase protocol that organizes a controlled session, and
how the resulting streams are stored.

\subsection{The Instrument Set}

This subsection is organised by proxy rather than by device: the instruments a
tier adds are the reading apparatus for the intent proxy that tier measures, and
the criterion for choosing one is single---whether it can physically reach the
quantity without altering the behaviour being measured. The three proxies sit
differently against that criterion, so the hardware differs, and the order of
depth follows from it.

  \textbf{L1, Attention---the $m_\text{gaze}$ proxy.} Visual orienting
  projects onto the external scene as light, so a headband that never touches the
  operator can record the scene from which gaze is inferred. L1 is not a single
  device but a collection layer assembled from fourteen third-party headbands
  supplied by thirteen vendors. All fourteen provide a
  forward stereo pair (two of them as a side-by-side image split into left and
  right, and one through a pair of its positioning cameras rather than a
  dedicated front pair). Seven add fisheye or side views for peripheral
  coverage.
  Native resolutions range from $512\times512$ (rectified) to
  $3840\times1200$, and video rates from 20 to 60\,Hz;
  \cref{tab:l1units} lists the units.

  \textbf{L2, Exertion---the $m_\text{emg}$ proxy.} Action potentials raised
  at the muscle fibres reach the skin surface by volume conduction, so eight
  electrodes per forearm over the flexor and extensor bellies suffice, sampled at
  2000\,Hz. The cost is contact: impedance degrades with sweat and movement. From
  L2 onward the headband is fixed at four lenses at 1920$\times$1200, which
  distinguishes it from the heterogeneous L1 units.

  \textbf{L3, Cognition---the $m_\text{eeg}$ proxy.} Preparatory cortical
  activity crosses the skull only as a field potential, small in amplitude and
  sensitive to movement. The headband and the EMG are identical to L2, and two
  640$\times$480 wrist cameras at 30\,fps add the near-hand view the head cannot
  see. EEG comes in two configurations: an 8-channel cap connected through the
  headband computer, for the open setting in which the operator moves freely; and
  a 32- or 128-channel cap connected through the fixed bench host, for the
  controlled setting in which it is joined by the binocular eye tracker, the three
  6-DoF trackers, the 20-node gloves, and the third-person RGB-D view.

\begin{figure}[!t]
  \centering
  \includegraphics[width=\linewidth]{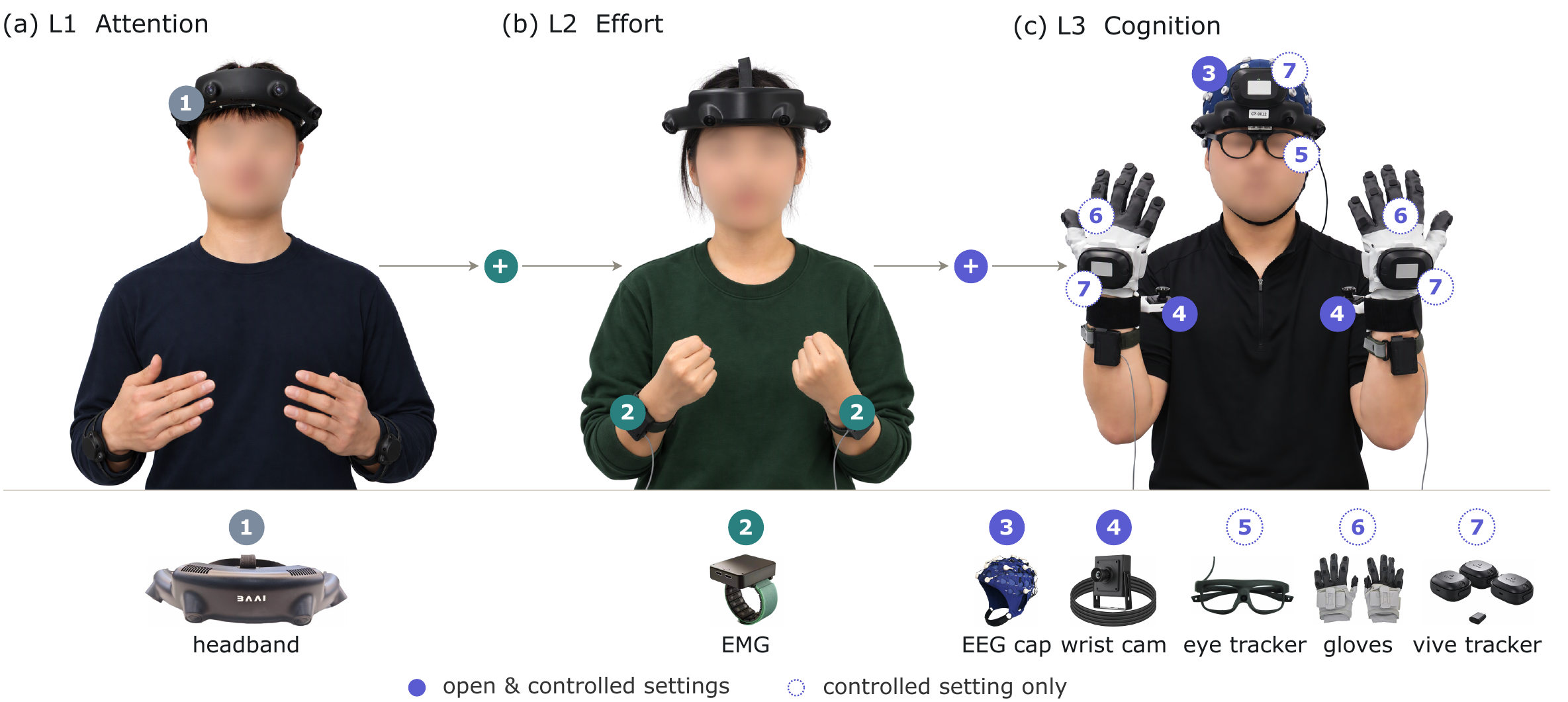}
  \caption{Instrumentation by tier. (a)~The L1 kit: a four-camera headband. (b)~The L2 kit adds two surface-EMG armbands worn on the forearm. (c)~The L3 kit adds the two wrist cameras and the EEG cap, and, in the controlled setting only, the eye tracker, the motion-capture gloves, and three 6-DoF trackers (head and both wrists). The devices each tier introduces stand in a row beneath its operator, numbered as on the body, and only the devices a tier adds are marked on that operator; solid numbers are hardware worn in both settings, dashed numbers the hardware used in the controlled setting only. Faces are blurred for anonymity.}
  \label{fig:hw}
\end{figure}

The correspondence between channel count and connection is not incidental: the
denser the cap, the higher the data rate and the tighter the electrode stability
required, and the more the record depends on a fixed station and a still posture.
Hence the distinction this subsection turns on. \emph{The same proxy is obtained
differently under different conditions}: gaze, head and wrist pose, and finger
kinematics are measured directly, with dedicated instruments, only within the
controlled L3 subset; at L1, at L2, and within the open L3 subset, the same
quantities are inferred from headband video. ``L3 as reference'' refers to
that controlled record. What the tiers add, instrument by instrument, is the
upgrade of an estimate into a measurement, and the preconditions of that upgrade
tighten with depth---which is why the deepest tier is also the narrowest.

Two invariants hold across the kit. First, no instrument is daisy-chained through
another: a single host receives every stream and stamps arrival offsets, so each
device keeps its own clock and native rate, and a failure or a dropped frame in
one stream cannot contaminate the others; that host is the headband computer in
the open setting and the bench host in the controlled one. Second, phase and run
transitions are driven through a parallel-port box into the EEG amplifier as well,
so protocol events are stamped on the amplifier's clock and not only on the
host's. Every camera is stored twice, raw and rectified---the depth view
rectified only---and the calibration ships with every episode. Poses share one frame throughout: $x$ points left, $y$ up, and
$z$ forward, with millimetres for translation and degrees for joint angles; the
alignment and calibration pipeline is described in
\cref{sec:sync}. \cref{fig:hw}(a)--(c) show the three kits in tier order: the headband at L1, one
armband on each forearm at L2, and at L3 the same kit with the EEG cap and a wrist
camera on each forearm. Only the devices a tier adds are numbered on its operator;
the eye tracker, the gloves and the head and wrist trackers are controlled-setting
only, which the dashed numbers mark.
The full modality matrix is in \appref{app:hw}, and the fourteen L1 units are
audited in \appref{app:l1audit}.

\subsection{Settings and Protocol}
\label{sec:protocol}

A tier's collection is governed by two requirements: how much of the everyday scene it must preserve, and how much signal fidelity its instrumentation demands. The two pull in opposite directions, and the corpus resolves them tier by tier.

\noindent\textbf{Two settings.} L1 and L2 privilege mobility over signal depth. Every instrument they add is wearable and connects to a headband-mounted computer, so the operator moves through everyday environments---standing, turning, walking, and gesturing---without a fixed bench or cable constraints. This is the \emph{open} setting: L1 and L2 are collected in it throughout, and L3 also uses it, with a lighter configuration, trading EEG density for scene coverage. Open here names the acquisition setting, not the task: an open-setting session can still use a scripted task sheet and a staged scene.

L3 is where the trade-off reverses. Dense electrophysiology requires a stable posture: electrode contact and impedance degrade with movement, and the denser the recording, the less it tolerates motion. The eye tracker, the gloves, and the wrist trackers load the same platform, so acquisition calls for a bench that does not move. Sessions run in short blocks, with a re-baseline between blocks to limit drift. L3 is collected under both settings: in the \emph{controlled} setting the amplifier and the host remain on a dedicated desk and the operator holds still, and the session runs imagery then execution; in the \emph{open} setting the kit runs on the headband computer in everyday scenes, and the session runs execution only.

\begin{wrapfigure}{r}{0.5\linewidth}
  \centering
  \includegraphics[width=\linewidth]{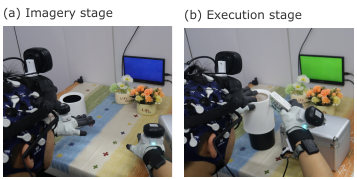}
  \caption{The two cued phases of a controlled session (e.g., \textit{pick up the cup}). (a)~Imagery (blue screen): the operator sits still, hands on the desk, and rehearses the task with only EEG and gaze recorded. (b)~Execution (green screen): the same operator manipulates the objects, head held as still as the grasp allows. The eyes move in both phases, the hands only in the second.}
  \label{fig:phases}
\end{wrapfigure}

\noindent\textbf{Imagery, then execution.} The controlled setting also carries the paradigm that makes the EEG informative, built around the high-density record. Imagery is confined to this setting for two reasons: its boundaries are delimited by phase markers that the experimenter records during the session, and the preparatory signal is low in amplitude and spatially structured, so only the dense array separates it from noise. Preparation and motor imagery are brief and are easily masked by the movement that follows, so the session separates imagery from execution (\cref{fig:phases}). In the imagery phase (\cref{fig:phases}a) the screen turns blue: the operator faces it, sits still, and rehearses the task sequence without moving, so that motor imagery and preparatory EEG are recorded free of movement artefact; gaze is recorded as well, so that the scanpath of the plan can be read off. In the execution phase (\cref{fig:phases}b) the screen turns green: the same operator manipulates the objects on the desk, and every modality is recorded, so that the EEG and the other streams can be read against the actual movement.
Every transition is also written to the record as a marker---run start, imagine start, execute start, and run end---so that the phases can be cut out of any stream on a common time base rather than by hand.

Each controlled episode therefore comprises three segments, not two cues alone: preparation while the operator settles, imagery of the task sequence, and execution of that sequence. Because the first two are acquired while the body is still, the EEG records the task as it is planned rather than as it is performed: the step sequence and the action each step will require, with no movement to mask them. The high-density record exists for that pairing: the same task as a plan, acquired while the body is still, and as a performance seconds later. Open settings (L1, L2, and L3 with the lighter configuration) record the execution segment alone.

\noindent\textbf{Collection procedure.} L1 is recorded without guidance: operators are asked only to carry out the activity as they normally would. L2 and L3 use task sheets: the operator arranges the objects as drawn and follows the step order marked on the sheet, so that the same task is staged identically across sessions. At L3, tasks are drawn in random order rather than blocked, which sustains engagement and limits drift of the EEG towards drowsiness. Operators also rotate between scenes at a fixed interval, so that no operator works a single scene for an entire shift and no scene is shaped by an individual's habits.

All operators are trained to a single standard before collection: how to wear each device, how to run a session, and how to record it. L3 adds two requirements: operators are trained to keep the head still during acquisition and are responsible for seating and maintaining the electrodes so that contact quality stays within range; participants assigned to L3 are also screened so that the EEG is clear and stable. Each hour of collection is followed by fifteen minutes of rest, to sustain attention and energy.

\subsection{Data Storage}
\label{sec:storage}

Streams are stored as recorded and linked by reference, never merged into a single file. Each device keeps its own clock, its own native sampling rate, and its own coordinate frame; forcing them into one stream would discard the timing the corpus exists to keep. Synchronization is therefore a property the data \emph{declares} rather than one it enforces: no resampling and no alignment happen at write time, and the corpus is written as many asynchronous files, each at its own rate.

\begin{figure}[!t]
  \centering
  \includegraphics[width=\linewidth]{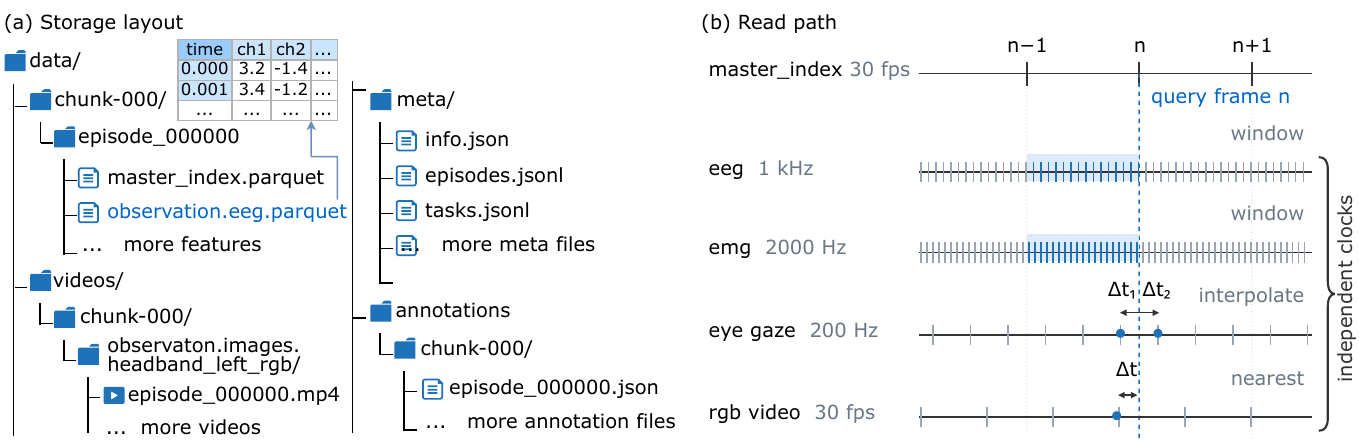}
  \caption{One episode on disk, and how it is read. (a)~One Parquet file per stream, each a timestamp column plus data columns, beside a master index. (b)~Timestamps are relative to the start of the recording, so no feature sits on a fixed grid; a query fixes a master rate and resolves each feature from its own timestamps. Every camera appears twice under \texttt{videos/}, once raw and once rectified, except the depth view, which is rectified only.}
  \label{fig:access}
\end{figure}

At write time, every episode keeps one Parquet table per modality (\cref{fig:access}a), one file per observation stream: the camera views, EEG, gaze, hand pose, glove skeleton, three device poses (head and both wrists), head and wrist IMUs, and wrist EMG. \texttt{videos/} holds one encoded video per camera view, \texttt{calibrations/} holds the rectification parameters and the manifest, \texttt{annotation/} the structured annotation and the QC verdict, and \texttt{meta/} the episode metadata with its alignment and statistics reports. Every sample is stamped relative to the start of the recording, and that difference is stored as given, not snapped to a master frame, so non-uniform intervals and device jitter survive into the file. The complete directory layout is in \appref{app:layout}.

Quality is enforced on the recording, not on the downstream analysis. Every session is checked stream by stream once it is written---dropped or out-of-order samples, black or frozen video, dead channels, impossible value jumps, a device missing from the set---and a session that fails is recaptured rather than repaired, so a released episode never mixes a good half with a patched one. The verdicts travel with the episode in its metadata; \appref{app:qc} lists the checks, their thresholds, and what each level costs.

Reading is fully decoupled from writing. A separate master-index file is sampled at a fixed rate, and the reader chooses that rate (30\,fps by default); a master frame then indexes every modality through its own timestamps, resolved by nearest sample, interpolation, or all samples in a window such as $(-1/\mathrm{FPS},\,0]$ (\cref{fig:access}b). The format is a modification of LeRobot~\cite{cadene2024lerobot} and remains compatible with it: installing \texttt{pip install humancoin} is enough to preview, download, and use the data, and the acquisition software and the documentation are released alongside it\footnote{\url{https://github.com/FlagOpen/HumanCOIN}}.

\section{Data Processing}
\label{sec:annotation}

\begin{figure}[!t]
  \centering
  \includegraphics[width=\linewidth]{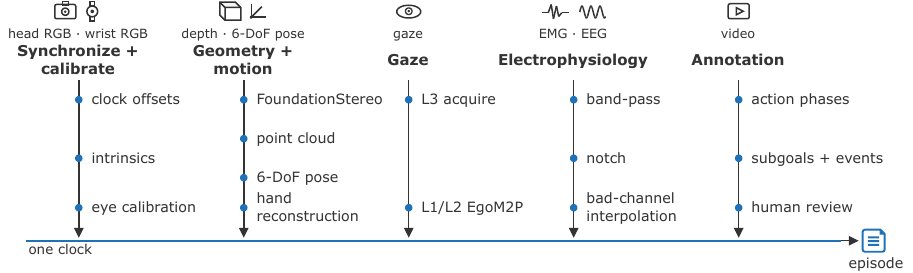}
  \caption{Processing pipeline. Each stream enters the group that handles its physics: synchronization and calibration (clock offsets, camera intrinsics, eye calibration); geometry and motion (stereo depth from FoundationStereo, point cloud, 6-DoF poses with finger joints, hand reconstruction from the rectified front view); gaze (acquisition at L3, prediction at L1 and L2); and electrophysiology (band-pass filtering, notch, bad-channel interpolation). The groups then align to a common time base and pass through annotation before an episode is written.}
  \label{fig:pipeline}
\end{figure}

Processing runs per stream, because the noise and the failure modes of each modality are set by its own measurement physics; the streams meet only at the end (\cref{fig:pipeline}). Each stream is reduced to a timestamp, a unit, a declared frame, and an alignment report, and the five groups follow the dependency order of the data---synchronization and calibration, geometry and motion, gaze, electrophysiology---with annotation last, since it consumes the others. Chains are fixed, their parameters are logged per episode, and a failing step is flagged rather than filled, so any run can be replayed.

\subsection{Synchronization and Calibration}
\label{sec:sync}

We use soft synchronization. The recorder stamps every sample relative to the start of the recording, and the sync host records arrival offsets for the streams that share a transport; the two together place each sample on a common time base without resampling. A per-stream audit of released L3 episodes is reported in \appref{app:sync}: the high-rate streams align below one sample period, while the 30\,Hz cameras are frame-quantized. Calibration (intrinsics and distortion for every camera, the tracker-to-frame transform, and the eye-tracker calibration) ships with the episode in \texttt{calibrations/}, alongside both encodings of
every camera, so a reader can re-derive the geometry rather than accept the
undistortion applied at write time. Because neither offsets nor calibrations are applied at write time, a reader can align at any custom rate and choose nearest sample, interpolation, or a window. Residuals between the chosen rate and the samples actually read are written to the per-episode alignment report, so a downstream analysis can audit the timing it inherited.

\subsection{Geometry and Motion}

Depth is built from RGB. The left-front and right-front lenses are hardware-timestamp synchronized, so every stereo pair comes from the same instant; the pair, together with the camera parameters, is passed to FoundationStereo~\cite{wen2025foundationstereo}, which returns a depth map stored with the intrinsics that produced it. Point clouds are formed per frame by projecting depth through those intrinsics and are subsampled to a fixed budget.

Motion follows the same route. In the controlled L3 subset three 6-DoF trackers give the head and both wrists directly ($\sim$800\,Hz, millimetres, one shared frame), and the gloves give 20 joint angles per hand ($\sim$800\,Hz, degrees). At L1, at L2, and in the open L3 subset both hands are reconstructed from video over each continuous segment: the model builds on ACE-Ego-Hand~\cite{liu2026aceegohand} and takes clean-latent features from Wan2.2-Fun-5B-Control~\cite{videoxfun2026,wan2025}, and a bidirectional spatiotemporal decoder returns, for every frame, the two wrists and the finger joints in the camera frame, MANO shape and articulation, wrist translation, the intrinsics it assumed, and two confidence fields per hand---whether it is present, and how much of it is visible.\footnote{The model was developed by the authors and is described in full in the accompanying report~\cite{innata2026}.}

Before reconstruction, the main view of every unit is rectified to a common pinhole domain, the intrinsics are updated after cropping and resizing, and the analytic rays are generated from the updated intrinsics; a unit's declared calibration is used where it exists, and the pinhole parameters are predicted from the features where it does not. The intrinsics released with the track are those of this domain, not those of the raw lens. Overlapping windows are fused at their source timestamps: rotations combine in rotation space, one shape per hand and one camera per segment are fixed, and the geometry is regenerated, so that shape and scale do not drift as the visible part of the hand changes---what a user reads is a hand that keeps its size through a grasp. The output is already three-dimensional in the camera frame, so only the wrist root has to be anchored: its pixel is looked up in the depth map and back-projected, and the finger joints, three-dimensional relative to that root, ride along. The depth map remains the metric reference for the scene and the objects. In the controlled L3 subset the same fields are measured rather than inferred, which is what makes that record the reference against which the reconstruction is scored. All poses (head, both hands, hand kinematics) and the environment point cloud are mapped into one unified coordinate frame.

\subsection{Gaze}

Gaze is obtained by acquisition and by estimation. Both routes write the same fields---a timestamp and a pixel coordinate---in the pixels of the image each route reads: the scene camera of the eye tracker in the controlled L3 subset, the rectified head view at L1, at L2, and in the open L3 subset.

In the controlled L3 subset the eye tracker is calibrated per participant and per session and logs at 200\,Hz; its coordinates are pixels of the scene-camera image, whose video runs at 30\,fps. The calibration that produced each record---the intrinsics and distortion of that camera, with the transform that carries a ray into the head frame---ships with the episode, so a pixel can be back-projected and placed in the head frame. Because L3 records a reference measurement, it also serves as the test bed for the estimator used at L1 and L2.

Outside that subset there is no eye tracker---at L1, at L2, and in the open L3 subset---and the gaze point is predicted by a pipeline built on EgoM2P~\cite{li2025egom2p} that runs from video to coordinates. It reads the unit's rectified left-front view, where the unit provides one---left not because it is more accurate, but because rectification, depth, hands, and annotation all use that view, so cross-modal alignment needs no extra transform---under the model's fixed input specification: a 2\,s window, 16 frames, and a $256\times256$ input. The gaze points of a window are predicted and then linearly resampled to the window's frame count, so that every frame carries exactly one row; a tail shorter than a window is covered by an additional window aligned to the end of the video, so that the prediction is as long as the video. The output is fixed at 30\,Hz and is not interpolated to L3's 200\,Hz: the model's positional encoding fixes 30 tokens per second, and interpolation would only manufacture precision. The difference ships with the column, and a user who wants another rate can resample. Coordinates are pixels of the video: the model works inside a fixed $256\times256$ centre crop, and the pipeline maps its output back, so that a released coordinate indexes the frame a user has rather than a position inside the model's input. The prediction is written as a table isomorphic to L3's but registered under a separate feature name, so that measurement and estimate cannot be confused.

\subsection{Electrophysiology}

Both streams are cleaned with a fixed chain. EMG is filtered in zero phase and reduced to an envelope. The trace is mirror-padded at both ends, band-passed between 20 and 450\,Hz with second- and fourth-order Butterworth sections applied forward and backward, and comb-notched at 50\,Hz with a 2\,Hz bandwidth, with periodic extension and period alignment at the edges so that the notch does not ring there. The cleaned trace is rectified and reduced to one envelope per channel, and the raw trace is kept beside it as \texttt{*\_emg\_raw}. \appref{app:ephys} lists the orders, cut-offs, and padding steps.

EEG passes through seven. The channel is de-meaned, band-passed between 0.5 and 100\,Hz with the high cut held below Nyquist, and notched at 50\,Hz, which keeps the delta-to-gamma range in which motor-imagery and error-related potentials live. Bad channels are flagged by five joint rules---flat, electrode contact, bridging between neighbours, decorrelation from the rest of the montage, and high-frequency noise---and a session passes only when fewer than 30\% of its channels are flagged. An infomax ICA fitted once on the clean segments of a session removes ocular components, selected by correlation with the frontal channels, and muscle components, selected by their power above 30\,Hz. Flagged channels are then interpolated by spherical spline over the retained ones, and the cleaned trace is re-referenced to their average. The stages are released as separate tables---\texttt{eeg\_raw} as acquired, \texttt{eeg} filtered, and \texttt{eeg\_repair} after interpolation and component removal. \appref{app:ephys} gives the rule thresholds, the component criteria, and the electrode positions that lack a conventional coordinate.

Motion cannot be removed from a recording in which the operator moves, and even the deepest tier is stationary only by protocol (\cref{sec:protocol}). The cleaned EEG is therefore signal plus residual motion noise: we release the metrics that bound it---impedance pass rate, share of interpolated channels, and share of samples marked artefactual---and leave the choice of a stricter rejection threshold to the user. EMG is recorded at 2000\,Hz in millivolts, eight channels per arm; EEG at 1000\,Hz over 128 saline channels or at 250\,Hz over 32 or 8 gel channels, in microvolts, with every channel named by brain region and its impedance logged in kilohms as a quality input.

\subsection{Annotation and Segmentation}

Every episode is annotated at the level of task steps (\cref{fig:anatomy}c): temporally bounded segments carrying the step sentence, the action it names, the hand in use, and every object the sentence reaches. Annotation is produced with Argus~\cite{argus2026}, an open-source harness for dense annotation of robot-learning data, running on top of Qwen3.8-27B~\cite{qwen2026qwen38}. Argus reasons jointly over video, state, and sensor evidence to produce dense labels: action phases, subgoals, key events, outcomes, operator mistakes and recoveries, and object-state changes, each with an execution and outcome score and with the evidence and confidence behind it. It exports the labels as JSON and as LeRobot language columns, and a human reviewer checks every export before it enters the release. Objects are not mere name strings: for L2 and L3 every object recorded in an episode's object list was measured for size and weight, and the measurement is linked to the object instances appearing in each trajectory. Collective references (``the cups'') are expanded to their member objects, and hand-in-use is read from wrist-tracker displacement inside the step window when no explicit label exists.

L1 needs a different route. Because passive collection does not intervene, a raw L1 recording is one continuous stream that lasts hours, and episodes are cut after the fact: frames and the available annotation are passed to a VLM, which proposes episode boundaries and step segments, and a human reviewer confirms or adjusts every boundary before release. This is why L1 episodes vary widely in length while controlled L3 sessions are scripted, and why an L1 boundary carries the uncertainty of a segmentation rather than the certainty of a cue.

\section{Discussion}
\label{sec:discussion}

An egocentric corpus pairs an observation with an action and so answers what happened; why the action took that form belongs to the intent that preceded it, which is often settled before execution. \dataset{} collects that intent, recording proxies for it at three signal depths. Whether it holds up depends on work we have not done; we state its limitations first, its openings second.

\subsection{Limitations}

\noindent\textbf{The premise is untested.} That intent sits between observation and action and carries information the observation does not is untested here: the corpus only makes it testable, and policy training, lead-time baselines, and cross-site quality gates lie outside its scope. Nor can the opposite be excluded: intent may be redundant with observation, since gaze is predictable from the scene and preparatory EEG may follow the kinematics. What fails then is not the concept but the proxies; the negative result turns ``is intent redundant?'' from a position into a measurement.

\noindent\textbf{Coverage and representativeness.} L3 accounts for \shareLthree\% of the hours (500 h); its controlled subset is about 350 h (2.8\%) and is the part concentrated on the desk, and electrode stability bounds how long the densest episodes can run. The scene and task mix is not a random sample of manipulation. L1 is assembled from fourteen heterogeneous units whose camera counts, resolutions, calibration, and declared frames differ (\appref{app:l1audit}), so how much of an L1 statistic reflects behaviour rather than the device mix cannot yet be separated.

\noindent\textbf{Annotation and comparability.} Step boundaries and object attributes are human judgements---L1 boundaries from automatic segmentation with human review, attributes measured only at L2 and L3. Hand motion is estimated at L1, at L2, and in the open L3 subset and measured in the controlled L3 subset, so the estimated part carries a reconstruction error. Comparability holds only in part: timestamps, units, frames, and calibration are released as each source declares them and the declarations are uneven, while quality metrics such as the impedance pass rate exist only at L3.

\noindent\textbf{Long tails.} The vocabularies are long-tailed: rare classes have few samples within a tier, cross-tier comparisons are unevenly powered, and the L3 share limits the deep tier's rare classes, so ``the deep tier is more concentrated'' holds only qualitatively.

\subsection{Research Directions}

None of these questions can be answered on $(o,a)$ data alone.

\noindent\textbf{Testing the mediation.} One test compares $p(a\mid o,m)$ with $p(a\mid o)$: if the proxies add nothing, the latter is no worse. Another is a cross-tier ablation; L1 and L3 share the task ontology and differ in depth, so an L1-trained model applied to L3 mixes distribution shift with the proxy difference. The imagery phase supplies a third: it records a plan without its execution, so a model can be asked to read the plan without action evidence.

\noindent\textbf{Evaluating before the fact.} With only $o$ and $a$ a corpus scores after the fact; with intent one can ask before: how the action would change under a different intent, whether the decision stays readable when no action occurs, and whether a model predicts the action or the intent behind it. Such answers do not depend on success, so they retain signal where failures are rare.

\noindent\textbf{Transport.} Three depths, several sites, many operators, and fourteen units make transport measurable: which proxies stay calibrated across tiers, which labels stay stable across annotators, and whether any component is independent of the sensor.

\noindent\textbf{Applications.} Intent-conditioned policies, earlier failure prediction, corrections with less supervision, and an intent interface for human--robot collaboration follow once $z$ enters the model: the pay-off of the chain, not the test of whether it holds.

The corpus succeeds if it makes these questions falsifiable; a negative answer counts as much as a positive one.

\section{Conclusion}

An egocentric manipulation corpus usually records two things: the observation $o$ and the action $a$. It records what happened, not what was decided---and the decision is what the action is answerable to. \dataset{} takes that middle term, the intent $z$, as the object of collection: one task ontology and one time base, with proxies for $z$ recorded at three signal depths---gaze for attention, EMG for execution, EEG for decision---and three tiers nested in instrumentation but not in distribution: passive and open at L1, active and open at L2, active and controlled at L3. What results is an instrument for testing intent rather than a larger behavioural dataset: task families that recur across tiers observed at several depths, which makes three questions measurable---where intent becomes readable, whether it mediates the action beyond the observation, and whether it transports across tiers.

We do not claim that these questions are answered; the value of the corpus is that they can be falsified, including by the negative result that intent is redundant with observation. The data, the acquisition software, and the documentation are released openly for research use.

\section{Data Statement}
\label{sec:data}

All participants gave written informed consent before any recording and could withdraw at any time without penalty. Episodes are released only in de-identified form: no personally identifying information is included, faces are blurred in every released view, and all tiers---L1, L2, and L3---share the same consent and de-identification protocol. Task protocols and instrumentation differ (\cref{sec:protocol}): L1 is unguided, L2 and L3 use task sheets; L2 adds EMG, and L3 adds EEG and, in the controlled setting, eye tracking and motion capture, all de-identified the same way. Code is released under Apache-2.0 and data under CC BY-NC 4.0.

\FloatBarrier
\begingroup
  \interlinepenalty=10000
  \setlength{\bibsep}{2pt}
  \reportreferences{main}
\endgroup
\reportappendix
\clearpage
\section{Scene Classes}
\label{app:scenes}
\begin{figure}[!t]
  \centering
  \includegraphics[width=\textwidth]{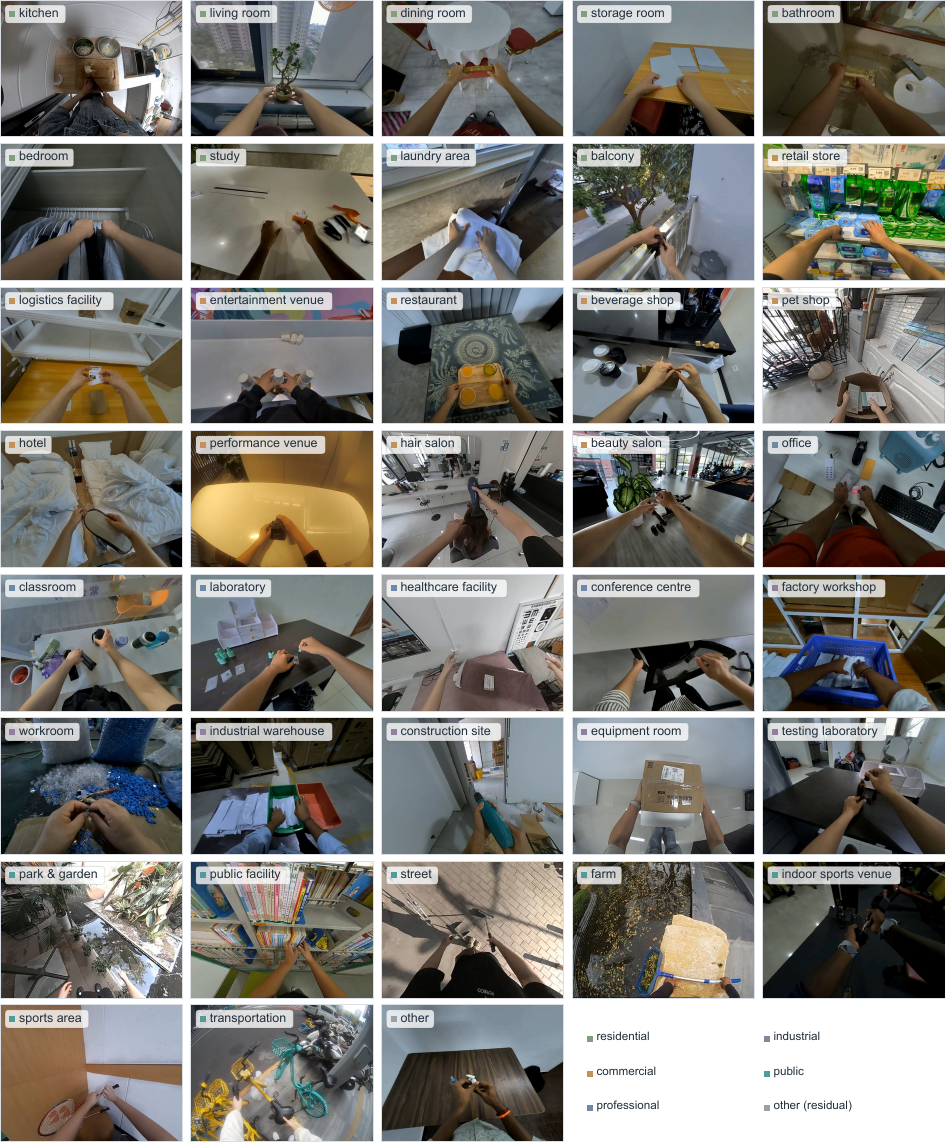}
  \captionof{figure}{Head-camera frames from the scene vocabulary: one frame for each of the \numscenes{} classes, labelled with its class name and ordered by zone (square colour) and by recorded hours within a zone. Per-class hours and tier coverage are in \cref{tab:scenes}.}
  \label{fig:scenes}
\end{figure}

The corpus spans \numscenes{} scene classes in \numzones{} zones---residential (\zoneResidential{} classes), commercial (\zoneCommercial), professional (\zoneProfessional), industrial (\zoneIndustrial), and public (\zonePublic)---plus one residual class. \cref{tab:scenes} lists all \numscenes{} with their hours in each tier, their share of the corpus, and the tiers in which they appear; \cref{fig:scenes} shows a recorded frame for each.

\begin{table}[H]
  \centering
  \caption{Scene classes in full. Hours are recorded hours by tier; total is the class total over all tiers; share is the percentage of the \numhours\,h corpus; tier gives the tiers in which the class appears. Zone headings give the class count, the zone's hours and share, and the number of classes in L3. Values are rounded to 0.1\,h, so rows and columns may not sum exactly.}
  \label{tab:scenes}
  {\footnotesize
  \setlength{\tabcolsep}{4pt}
  \renewcommand{\arraystretch}{0.86}
  \begin{tabular*}{\textwidth}{@{\extracolsep{\fill}}r >{\raggedright\arraybackslash}p{1.15in} r r r r r >{\raggedright\arraybackslash}p{0.62in}@{}}
  \hline
  \# & Class & L1 (h) & L2 (h) & L3 (h) & Total (h) & Share (\%) & Tier \\
  \hline
  \multicolumn{8}{@{}l}{\emph{Residential --- 9 classes, 4,520.7\,h, 36.2\% of the corpus, 4 in L3}}\\
  \hline
  1 & kitchen & 868.2 & 271.5 & 35.1 & 1,174.8 & 9.4 & L1--L3 \\
  2 & living room & 759.0 & 43.3 & 0.0 & 802.2 & 6.4 & L1--L2 \\
  3 & dining room & 400.5 & 174.1 & 75.2 & 649.9 & 5.2 & L1--L3 \\
  4 & storage room & 502.8 & 98.4 & 0.0 & 601.2 & 4.8 & L1--L2 \\
  5 & bathroom & 338.9 & 169.2 & 0.0 & 508.1 & 4.1 & L1--L2 \\
  6 & bedroom & 431.3 & 51.2 & 0.0 & 482.5 & 3.9 & L1--L2 \\
  7 & study & 95.4 & 34.0 & 34.5 & 163.9 & 1.3 & L1--L3 \\
  8 & laundry area & 69.4 & 25.0 & 30.7 & 125.1 & 1.0 & L1--L3 \\
  9 & balcony & 13.0 & 0.0 & 0.0 & 13.0 & 0.1 & L1 \\
  \hline
  \multicolumn{8}{@{}l}{\emph{Commercial --- 10 classes, 2,588.1\,h, 20.7\% of the corpus, 2 in L3}}\\
  \hline
  10 & retail store & 355.0 & 157.4 & 32.7 & 545.1 & 4.4 & L1--L3 \\
  11 & logistics facility & 401.0 & 127.9 & 0.0 & 528.9 & 4.2 & L1--L2 \\
  12 & entertainment venue & 472.8 & 0.0 & 0.0 & 472.8 & 3.8 & L1 \\
  13 & restaurant & 197.7 & 28.0 & 0.0 & 225.7 & 1.8 & L1--L2 \\
  14 & beverage shop & 141.2 & 32.0 & 29.7 & 202.9 & 1.6 & L1--L3 \\
  15 & pet shop & 202.6 & 0.0 & 0.0 & 202.6 & 1.6 & L1 \\
  16 & hotel & 113.0 & 38.4 & 0.0 & 151.3 & 1.2 & L1--L2 \\
  17 & performance venue & 123.8 & 0.0 & 0.0 & 123.8 & 1.0 & L1 \\
  18 & hair salon & 78.4 & 0.0 & 0.0 & 78.4 & 0.6 & L1 \\
  19 & beauty salon & 56.5 & 0.0 & 0.0 & 56.5 & 0.5 & L1 \\
  \hline
  \multicolumn{8}{@{}l}{\emph{Professional --- 5 classes, 1,982.3\,h, 15.9\% of the corpus, 2 in L3}}\\
  \hline
  20 & office & 789.2 & 177.1 & 162.4 & 1,128.6 & 9.0 & L1--L3 \\
  21 & classroom & 338.9 & 12.5 & 0.0 & 351.4 & 2.8 & L1--L2 \\
  22 & laboratory & 215.7 & 37.5 & 34.1 & 287.3 & 2.3 & L1--L3 \\
  23 & healthcare facility & 200.3 & 0.0 & 0.0 & 200.3 & 1.6 & L1 \\
  24 & conference centre & 14.7 & 0.0 & 0.0 & 14.7 & 0.1 & L1 \\
  \hline
  \multicolumn{8}{@{}l}{\emph{Industrial --- 6 classes, 1,956.7\,h, 15.7\% of the corpus, 1 in L3}}\\
  \hline
  25 & factory workshop & 857.1 & 311.1 & 0.0 & 1,168.2 & 9.3 & L1--L2 \\
  26 & workroom & 387.6 & 110.0 & 65.5 & 563.2 & 4.5 & L1--L3 \\
  27 & industrial warehouse & 126.6 & 0.0 & 0.0 & 126.6 & 1.0 & L1 \\
  28 & construction site & 74.5 & 0.0 & 0.0 & 74.5 & 0.6 & L1 \\
  29 & equipment room & 15.6 & 0.0 & 0.0 & 15.6 & 0.1 & L1 \\
  30 & testing laboratory & 8.7 & 0.0 & 0.0 & 8.7 & 0.1 & L1 \\
  \hline
  \multicolumn{8}{@{}l}{\emph{Public --- 7 classes, 1,210.6\,h, 9.7\% of the corpus, 0 in L3}}\\
  \hline
  31 & park \& garden & 247.6 & 17.7 & 0.0 & 265.3 & 2.1 & L1--L2 \\
  32 & public facility & 215.3 & 0.0 & 0.0 & 215.3 & 1.7 & L1 \\
  33 & street & 193.7 & 0.0 & 0.0 & 193.7 & 1.5 & L1 \\
  34 & farm & 150.7 & 19.7 & 0.0 & 170.4 & 1.4 & L1--L2 \\
  35 & indoor sports venue & 157.6 & 0.0 & 0.0 & 157.6 & 1.3 & L1 \\
  36 & sports area & 118.4 & 0.0 & 0.0 & 118.4 & 0.9 & L1 \\
  37 & transportation & 90.0 & 0.0 & 0.0 & 90.0 & 0.7 & L1 \\
  \hline
  \multicolumn{8}{@{}l}{\emph{Other --- 1 class, 241.6\,h, 1.9\% of the corpus, 0 in L3}}\\
  \hline
  38 & other & 177.5 & 64.1 & 0.0 & 241.6 & 1.9 & L1--L2 \\
  \hline
  \end{tabular*}}
\end{table}

\noindent\textbf{What a class is.} A class names the physical setting in which an episode was recorded, not the activity performed there: the same task may occur in several classes, and a class hosts many tasks. Every episode carries a single scene label, so the class hours partition the corpus and sum to \numhours. The five zones are functional strata for reporting and stratified use, not a hierarchy---they group settings by primary function (living, commerce, work, production, and public life)---and two classes in one zone need not be closer than two in different zones. The residual class is a label like the others, not missing data: it holds 241.6\,h, 1.9\% of the corpus, and appears at L1 and L2 but not L3.

\noindent\textbf{Boundaries.} The label follows the setting in which the hands act for the greater part of the episode; crossing a doorway does not change it. The pairs that carry most of the ambiguity are resolved by function, not by furniture: kitchen against dining room separates preparation and cleanup from eating and serving; office against study separates a workplace from a home workspace, so the same desk task is labelled by the setting; storage room against industrial warehouse separates domestic from industrial storage, and hotel against bedroom a commercial guest room from a home bedroom. In the public zone, street is outdoor circulation, park and garden managed green space, and public facility an indoor public building; restaurant is a sit-down meal and a beverage shop a drinks counter.

\noindent\textbf{The nine L3 classes.} The deep tier does not spread evenly over the vocabulary. Office (162.4\,h), dining room (75.2), workroom (65.5), and kitchen (35.1) carry 67.6\% of its 500\,h; the other five are study (34.5), laboratory (34.1), retail store (32.7), laundry area (30.7), and beverage shop (29.7). The nine span four zones---residential, commercial, professional, and industrial---and no public-zone class reaches L3. The set is therefore concentrated but not confined to one room type: two of its nine classes are shops.

\section{Verb Lexicon}
\label{app:verbs}
Every step carries two labels rather than one. A \emph{task} verb names the goal the step accomplishes and is the unit a plan is written in; the \emph{motion verbs} recorded inside the step name what the hand does to reach that goal, 2.5 of them per step. The two vocabularies are disjoint---no verb appears in both layers---and they behave in opposite ways: the task layer is open and long-tailed, the motion layer closed and concentrated. \cref{tab:verbs} lists both in full, with a boundary statement for each verb, the tiers in which it is observed, and its total occurrences; per-tier counts ship with the release.

\noindent\textbf{The task layer.} \numsubverbs{} verbs over the annotated steps. The ten most frequent take \pctSubverbsTop\% of the record and the median verb occurs \medSubverbClass{} times, but the tail is long: 28 verbs occur 25 times or fewer, down to \emph{uncap} at 4. Coverage varies by tier---98 verbs are observed at L1, 79 at L2 and \numsubverbsLthree{} at L3; 34 appear in all three and 15 in only one. What a protocol removes is informative. Fifty-nine verbs are absent at L3, and the ones that matter for the protocol fall into three groups: four are cleaning (\emph{spray}, \emph{rinse}, \emph{wash}, \emph{dry}), six are cord or closure work (\emph{seal}, \emph{stick}, \emph{thread}, \emph{untie}, \emph{coil}, \emph{connect}), and four are whole-object actions (\emph{throw}, \emph{unfold}, \emph{cover}, \emph{peel}); \emph{tear}, which never reaches L2, is absent for a simpler reason. A scripted desk task does not throw, unfold or wash.

\noindent\textbf{The motion layer.} \nummotionverbs{} verbs over the motion annotations, and effectively a closed set: the four commonest---\emph{place}, \emph{grasp}, \emph{pick}, \emph{move}---take 82.2\% of all motion annotations and the ten most frequent take \pctMotionverbsTop\%, against 35.8\% and \pctSubverbsTop\% for the same ranks in the task layer. Every one of the \nummotionverbs{} verbs occurs at L1. Twenty-four occur in all three tiers, one (\emph{tilt}) at L1 and L3 only, and 21 are absent at L3, falling into three groups: all six fine-contact verbs (\emph{pinch}, \emph{poke}, \emph{feel}, \emph{rub}, \emph{tap}, \emph{squeeze}), all five transport verbs (\emph{carry}, \emph{drag}, \emph{slide}, \emph{transfer}, \emph{bring}), and ten posture or alignment verbs (\emph{push}, \emph{lay}, \emph{spread}, \emph{roll}, \emph{stand}, \emph{dip}, \emph{lean}, \emph{bend}, \emph{extend}, \emph{retract}). What drops out is contact without motion, and transport over distance.

\noindent\textbf{The two layers together.} The task verb is the sentence a plan is written in and the motion verbs are the record of how the hand executed it, and the tiers treat the two differently: L3 keeps \numsubverbsLthree{} of \numsubverbs{} task verbs (40\%) but \nummotionverbsLthree{} of \nummotionverbs{} motion verbs (54\%), and the motion verbs it keeps carry \pctMotionverbsLthreeOfLone\% of L1's motion annotations against \pctSubverbsLthreeOfLone\% for the task verbs. The decomposition is also finer where the record is deepest---2.50 motion verbs per step at L1, 3.00 at L3---so the deep tier says less about what was done and more about how.

{\small
\setlength{\tabcolsep}{4pt}
\begin{longtable}{@{}r p{0.95in} p{3.95in} c r@{}}
\caption{The verb lexicon in full. A \emph{task} verb names the goal a step accomplishes; a \emph{motion} verb names what the hand does inside the step. Glosses state the boundary used by the annotators, tiers give where the verb is observed, and share is each verb's percentage of the annotations in its layer; percentages may not sum to 100 because of rounding.}\label{tab:verbs}\\
\hline
\# & Verb & Gloss & Tiers & Share (\%) \\
\hline
\endfirsthead
\hline
\# & Verb & Gloss & Tiers & Share (\%) \\
\hline
\endhead
\hline
\endfoot
\multicolumn{5}{@{}l}{\emph{Task layer --- 99 verbs; share of all task annotations}}\\
\hline
1 & wipe & clean a surface with a cloth kept in contact & L1--L3 & 12.39 \\
2 & open & move a closure so the container or space becomes accessible & L1--L3 & 9.93 \\
3 & fold & bend a flexible object back onto itself & L1--L3 & 7.75 \\
4 & close & return a closure to its shut position & L1--L3 & 5.75 \\
5 & arrange & set several objects into an intended spatial order & L1--L3 & 5.47 \\
6 & remove & take an object away from its current place & L1--L3 & 5.26 \\
7 & throw & release an object so it travels through the air & L1, L2 & 4.87 \\
8 & organize & group objects into an intended arrangement or container & L1--L3 & 3.26 \\
9 & pour & tip a vessel so its contents fall into a target & L1--L3 & 2.75 \\
10 & unfold & open out something previously folded & L1, L2 & 2.56 \\
11 & collect & gather scattered objects into one place & L1--L3 & 2.55 \\
12 & hang & suspend an object from a support & L1--L3 & 2.41 \\
13 & insert & place a part into a receptacle and seat it & L1--L3 & 2.13 \\
14 & stack & place objects one on top of another & L1--L3 & 2.08 \\
15 & clean & remove dirt from a surface or object & L1--L3 & 2.02 \\
16 & cover & place something over an object to hide or protect it & L1, L2 & 1.82 \\
17 & wrap & enclose an object in a flexible sheet & L1--L3 & 1.34 \\
18 & cut & divide an object with a blade & L1--L3 & 1.32 \\
19 & secure & fix an object so it cannot move & L1--L3 & 1.31 \\
20 & tear & divide a flexible object by pulling it apart & L1 & 1.20 \\
21 & install & mount a unit in its working position and fix it there & L1--L3 & 1.17 \\
22 & spray & release a liquid from a nozzle over a target & L1, L2 & 1.13 \\
23 & rinse & run water over an object to carry residue away & L1, L2 & 1.11 \\
24 & sort & distribute objects among classes by an attribute & L1--L3 & 1.02 \\
25 & tie & fasten with a cord drawn into a knot & L1--L3 & 0.97 \\
26 & tighten & turn a fastener until the joint stops yielding & L1--L3 & 0.94 \\
27 & sweep & push loose material along a surface with a brush & L1--L3 & 0.91 \\
28 & attach & join a part to a host so that the two move together & L1--L3 & 0.90 \\
29 & seal & close an opening so that it no longer passes air or liquid & L1, L2 & 0.88 \\
30 & unscrew & turn a threaded fastener out of its seat & L1--L3 & 0.84 \\
31 & stick & fix a flat item to a surface by adhesion & L1, L2 & 0.75 \\
32 & coil & wind a flexible line into concentric loops & L1, L2 & 0.67 \\
33 & screw & turn a threaded fastener into its seat & L1--L3 & 0.58 \\
34 & disassemble & separate an assembly into its parts & L1--L3 & 0.56 \\
35 & wash & clean an object with water and agitation & L1, L2 & 0.53 \\
36 & check & inspect an object against an expected state & L1--L3 & 0.52 \\
37 & assemble & join parts into a working whole & L1--L3 & 0.51 \\
38 & connect & join two ends so that a link passes between them & L1, L2 & 0.47 \\
39 & replace & put a new part where the old one was taken out & L1--L3 & 0.36 \\
40 & dry & remove moisture from a surface by wiping or by air & L1, L2 & 0.36 \\
41 & thread & pass a line through an eye or an opening & L1, L2 & 0.32 \\
42 & write & deposit marks on a surface with a held implement & L1--L3 & 0.32 \\
43 & untie & undo a knot and free the cord & L1, L2 & 0.27 \\
44 & peel & strip an outer layer from an object by hand & L1, L2 & 0.25 \\
45 & rearrange & move objects already placed into a new order & L1--L3 & 0.25 \\
46 & button & fasten a garment through its buttonhole & L1, L2 & 0.24 \\
47 & separate & part two joined objects so that they no longer touch & L1--L3 & 0.23 \\
48 & apply & spread a substance over a surface & L1--L3 & 0.22 \\
49 & scrape & draw a hard edge along a surface to lift material off & L1, L2 & 0.22 \\
50 & cap & fit a cover onto an opening & L1--L3 & 0.22 \\
51 & untangle & separate a snarl of flexible material into free ends & L1, L3 & 0.21 \\
52 & reassemble & rejoin the parts of an assembly taken apart earlier & L1, L2 & 0.21 \\
53 & scrub & clean a surface with a brush under pressure & L1, L2 & 0.18 \\
54 & scoop & lift loose material with a cupped tool & L1 & 0.17 \\
55 & bind & fasten several objects together as one bundle & L1, L2 & 0.17 \\
56 & wind & turn a flexible line around an axis & L1 & 0.17 \\
57 & unzip & open a zip fastener & L1, L2 & 0.17 \\
58 & slice & divide food into thin flat pieces & L1, L2 & 0.16 \\
59 & crumple & compress a sheet into an irregular ball & L1 & 0.16 \\
60 & staple & fix sheets together with a metal staple & L1, L2 & 0.15 \\
61 & zip & close a zip fastener & L1, L2 & 0.15 \\
62 & draw & trace a line on a surface with a held implement & L1, L3 & 0.15 \\
63 & fill & put material into a container up to a level & L1, L2 & 0.13 \\
64 & erase & remove marks from a surface by rubbing & L1, L3 & 0.11 \\
65 & unplug & withdraw a plug from its socket & L1, L2 & 0.11 \\
66 & test & operate an object to see whether it works & L1--L3 & 0.10 \\
67 & stamp & press an inked face onto a surface & L1 & 0.09 \\
68 & wring & twist a wet cloth to press the water out & L1, L2 & 0.08 \\
69 & loosen & turn a fastener until it no longer holds & L1 & 0.08 \\
70 & stir & move a tool through a mixture to blend it & L1, L3 & 0.08 \\
71 & drape & lay a flexible sheet over a form so that it hangs & L1 & 0.07 \\
72 & load & put an object into a carrier for transport & L1, L2 & 0.07 \\
73 & discard & put an object into waste & L1, L3 & 0.06 \\
74 & tuck & push an edge of flexible material under or into & L1 & 0.06 \\
75 & unwrap & remove a sheet that encloses an object & L1 & 0.06 \\
76 & detach & take a part off its host & L1, L2 & 0.06 \\
77 & trim & cut a small excess from an edge & L1, L2 & 0.06 \\
78 & divide & split one quantity into parts & L1, L2 & 0.06 \\
79 & unroll & unroll a wound sheet or line & L1, L2 & 0.06 \\
80 & punch & make a hole by pressing a tool through & L1 & 0.06 \\
81 & reattach & attach a part again after it was detached & L1, L2 & 0.06 \\
82 & scan & pass a reader over a code to register it & L1 & 0.05 \\
83 & plug & insert a plug into its socket & L1, L2 & 0.05 \\
84 & tape & fix something with adhesive tape & L1, L2 & 0.05 \\
85 & disconnect & undo a link between two ends & L1, L2 & 0.04 \\
86 & switch & change the state of a control & L1, L2 & 0.04 \\
87 & wet & apply water to an object & L1, L2 & 0.04 \\
88 & pry & lever an object open with a rigid tool & L1, L2 & 0.04 \\
89 & reinsert & insert a part again after removal & L1, L2 & 0.03 \\
90 & clamp & hold an object with a device that grips it & L1, L2 & 0.03 \\
91 & dispense & release a measured amount from a container & L1, L2 & 0.03 \\
92 & pack & put objects into a container for storage & L1, L2 & 0.03 \\
93 & brush & clean a surface with a brush & L1, L2 & 0.03 \\
94 & pair & bring two items together as a matched set & L1 & 0.02 \\
95 & polish & rub a surface until it shines & L1, L2 & 0.02 \\
96 & soak & leave an object in liquid for a period & L1, L2 & 0.02 \\
97 & dice & cut food into small cubes & L1 & 0.02 \\
98 & sew & join material with needle and thread & L1 & 0.02 \\
99 & uncap & take a cap off its opening & L3 & 0.01 \\
\hline
\multicolumn{5}{@{}l}{\emph{Motion layer --- 46 verbs; share of all motion annotations}}\\
\hline
1 & place & put an object down at a target pose & L1--L3 & 26.75 \\
2 & grasp & close the hand on an object so that it can be held & L1--L3 & 20.75 \\
3 & pick & lift an object off its support & L1--L3 & 19.99 \\
4 & move & change the position of a held object without releasing it & L1--L3 & 14.67 \\
5 & reach & carry the empty hand toward a target & L1--L3 & 3.26 \\
6 & push & move an object away by force applied through contact & L1, L2 & 1.88 \\
7 & adjust & nudge an object to correct its pose & L1--L3 & 1.83 \\
8 & transport & carry a held object between places & L1--L3 & 1.21 \\
9 & turn & rotate a held object about a vertical axis & L1--L3 & 1.12 \\
10 & pull & draw an object toward the body by force & L1--L3 & 1.02 \\
11 & lift & raise an object against gravity & L1--L3 & 0.80 \\
12 & hold & keep an object in the hand without moving it & L1--L3 & 0.70 \\
13 & release & open the hand so that the object leaves it & L1--L3 & 0.69 \\
14 & spread & lay something out over an area & L1, L2 & 0.67 \\
15 & roll & turn an object over on its side & L1, L2 & 0.54 \\
16 & lay & set an object down flat on a surface & L1, L2 & 0.53 \\
17 & align & bring an object into line with a reference & L1--L3 & 0.49 \\
18 & flip & turn an object over to its other face & L1--L3 & 0.47 \\
19 & straighten & remove a bend from a flexible object & L1--L3 & 0.38 \\
20 & press & push down on a surface through a small area & L1--L3 & 0.22 \\
21 & squeeze & compress a soft object in the hand & L1, L2 & 0.22 \\
22 & position & bring an object to a specified place without releasing it & L1--L3 & 0.21 \\
23 & reposition & move an object already placed to a new place & L1--L3 & 0.20 \\
24 & carry & move an object while supporting its weight & L1, L2 & 0.18 \\
25 & stand & set an object upright on its base & L1 & 0.18 \\
26 & dip & lower an object into a liquid and take it out again & L1, L2 & 0.16 \\
27 & rotate & turn an object about its own axis & L1--L3 & 0.15 \\
28 & shake & move an object to and fro rapidly & L1--L3 & 0.12 \\
29 & drop & let an object fall from the hand & L1--L3 & 0.11 \\
30 & bring & move an object toward a person or target & L1, L2 & 0.06 \\
31 & drag & move an object along a surface without lifting it & L1, L2 & 0.05 \\
32 & lower & move an object downward & L1--L3 & 0.04 \\
33 & tilt & incline an object from its upright pose & L1, L3 & 0.04 \\
34 & slide & move an object along a surface in contact with it & L1, L2 & 0.04 \\
35 & retract & draw the hand back from an object & L1, L2 & 0.03 \\
36 & tap & strike a surface briefly with the fingers & L1, L2 & 0.03 \\
37 & transfer & move an object from one hand or place to another & L1, L2 & 0.03 \\
38 & orient & turn an object so that a chosen face points at a target & L1--L3 & 0.03 \\
39 & lean & rest an object against a support at an angle & L1 & 0.03 \\
40 & adjust\_grip & change the placement of the fingers on a held object & L1--L3 & 0.03 \\
41 & rub & move the hand back and forth over a surface & L1, L2 & 0.02 \\
42 & bend & deform an object along its length & L1 & 0.02 \\
43 & pinch & grip a small object between two fingers & L1, L2 & 0.02 \\
44 & poke & push a finger into an object or an opening & L1, L2 & 0.02 \\
45 & extend & straighten the arm or the fingers outward & L1 & 0.01 \\
46 & feel & touch an object to judge its state & L1 & 0.01 \\
\end{longtable}
}

\section{Object Library}
\label{app:objects}

Every object an episode touches is an entry of a measured library: 839 instances under 442 distinct names, of which 829 are active and 10 retired, the retired ones shipped with a flag. An entry records a name, a shape together with the parameters that shape calls for, a primary and an optional secondary colour in hexadecimal, a material, a weight, a photograph, and a UUID. The UUID is the only key that joins an entry to the object list written with an episode, so the per-instance table ships with the release and is not printed here. The library grew with collection---35 entry dates between September and November 2026---rather than being assembled in advance.

\begin{figure}[h]
  \centering
  \includegraphics[width=\textwidth]{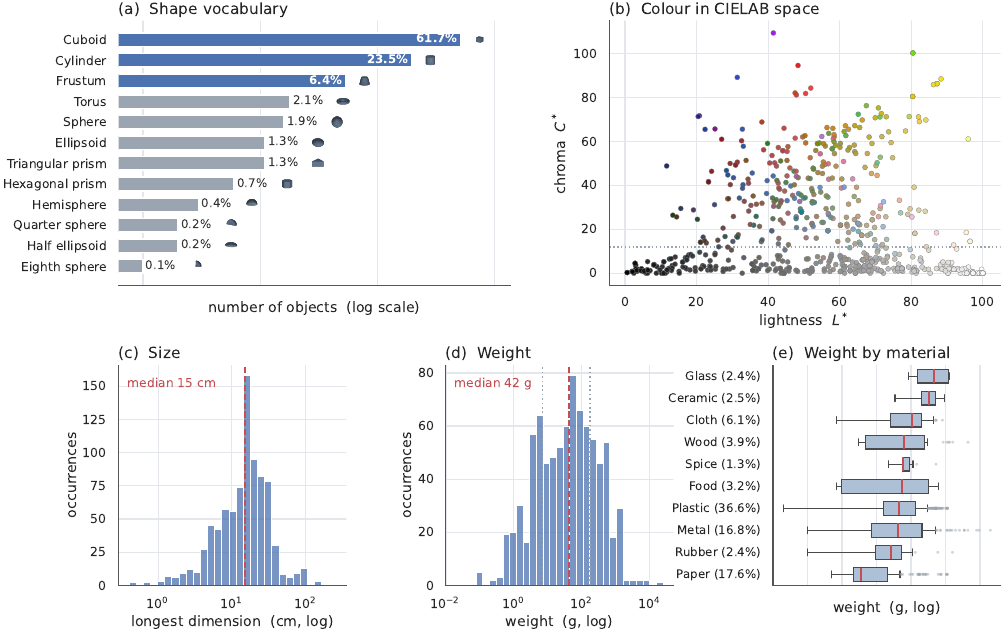}
  \caption{The object library. (a)~The twelve shape primitives and their percentage share of the library; the axis is logarithmic, and the solid beside each bar is the primitive. (b)~Colour of every instance in CIELAB, lightness against chroma; the dotted line at $C^*=12$ separates the achromatic entries. (c)~Longest dimension. (d)~Weight, with quartiles dotted and the median dashed. (e)~Weight of the ten commonest materials, ordered by median, with each material's percentage share in brackets. $n=839$.}
  \label{fig:objects}
\end{figure}

\noindent\textbf{Shape.} Twelve primitives are in use (\cref{fig:objects}a). Cuboid and cylinder alone account for 715 instances, or 85\%; the other ten cover 124, the largest being the frustum (54, the open vessels: cups, bowls, pots, bins) and the torus (18, objects with a through-hole: tape rolls, washers, a bracelet). Size is stored as the parameters the shape calls for---length, width, and thickness for a cuboid, diameter and height for a cylinder or a frustum, three axes for an ellipsoid---so one schema serves a mug, a box, and a sheet. Lengths are read with a caliper and weights on a scale, and both are written to one decimal place, 0.1\,cm and 0.1\,g, which is the resolution the table carries.

\noindent\textbf{Size.} By longest dimension the median instance is 15\,cm, with quartiles at 8.6 and 23\,cm. 86 instances are shorter than 5\,cm---screws, washers, nuts, the parts a hand manipulates at the fingertips---and 54 reach 40\,cm or more, the furniture and fabric at the other end. Between them, 84\% of the library falls in 5--40\,cm, the range in which a seated operator reaches and grasps (\cref{fig:objects}c).

\noindent\textbf{Weight and material.} Weight spans six orders of magnitude, from 0.02\,g (a plastic shaving) to 20\,kg (a shelving unit), with a median of 42\,g and quartiles at 7 and 172\,g. 29\% of the instances are under 10\,g and only 38, some 4.5\%, exceed 1\,kg: the library is one of small, light objects (\cref{fig:objects}d). Plastic accounts for 307 entries (37\%), paper for 148, metal for 141, together about seven in ten. Weight separates by material (\cref{fig:objects}e): glass and ceramic are highest, at medians of 470 and 326\,g, paper lowest at 3.6\,g, and plastic and metal both near 40--50\,g but with long tails. At a given size it is the material that decides whether an object can be lifted in one hand, which is why it is recorded. The three commonest shape--material pairs are cuboid--plastic (168), cuboid--paper (139), and cylinder--metal (50).

\noindent\textbf{Colour.} Colour is the dominant value sampled from the photograph, in hexadecimal, with an optional second value. Mapped to names the library is greyscale-dominated---grey 25\%, white 15\%, black 13\%, brown 11\%, transparent 11\%---and chromatic objects are only 35\% of the total. In CIELAB (\cref{fig:objects}b) most instances sit in the low-chroma band ($C^*<20$), and saturated reds and yellows are rare. Colour is therefore a weak discriminator between objects in one scene, which is why shape and size are recorded beside it.

\noindent\textbf{Counting.} The class count of \cref{sec:tasks} counts the object classes named by annotators in a step; this appendix counts the physical instances that were measured. Each measured instance carries a UUID that also appears in the object list written with every episode that used it; a class reaches the library only through the instances its steps name. The two counts are therefore not the same number: a class may be realised by several measured instances, and one instance may be named by more than one class.

\section{Hardware Bill of Materials}
\label{app:hw}
\begin{table}[h]
\centering
\caption{Stream specifications of the in-house kit, which records L2 and L3. Every camera records RGB and runs at 30\,fps; every view is released raw and rectified, the depth view rectified only; the third-person view exists only in the controlled setting. L1 records on third-party headbands, audited unit by unit in \appref{app:l1audit}. The pose and hand streams are released as fused outputs at about 800\,Hz; the trackers and the glove that feed them run at their own device rates.}
\label{tab:bom}
\small
\setlength{\tabcolsep}{2pt}%
\begin{tabular}{@{}>{\raggedright\arraybackslash}p{1.05in}>{\raggedright\arraybackslash}p{1.55in}>{\raggedright\arraybackslash}p{2.55in}>{\raggedright\arraybackslash}p{0.55in}>{\raggedright\arraybackslash}p{0.45in}@{}}
    \hline
    Stream & Device (model) & Resolution / channels & Rate & Unit \\
    \hline
    Head RGB $\times$4 & four-camera headband, Ego Headband V1.5 (in-house) & 1920$\times$1200 each; left-front, right-front, left, right; clock hardware-synced & 30\,fps & RGB \\
    Wrist RGB $\times$2 & JIERUIWEITONG FH868 & 640$\times$480 & 30\,fps & RGB \\
    Scene camera & Pupil Labs Neon & 1600$\times$1200 & 30\,fps & RGB \\
    Third-person view & Intel RealSense D455f & 640$\times$480 & 30\,fps & RGB \\
    Gaze & Pupil Labs Neon & $x,y$ pixel in the scene-camera image (L3), in the rectified head view (L1--L2) & 200\,Hz & px \\
    Head / wrist pose $\times$3 & HTC VIVE Tracker 3.0 & one shared frame: $+x$ left, $+y$ up, $+z$ forward & $\sim$800\,Hz & mm \\
    Finger kinematics & MANUS Metaglove Pro Haptics & 20 joint angles per hand & $\sim$800\,Hz & deg \\
    Surface EMG & WLEMGUV2A armbands $\times$2 & 8 channels per arm & 2000\,Hz & mV \\
    EEG (saline) & Neuroscan SynAmps2 (Curry) & 128 channels, brain-region labels, impedance logged & 1000\,Hz & $\mu$V \\
    EEG (gel) & BrainCo BCIGo & 32 or 8 channels, brain-region labels, impedance logged & 250\,Hz & $\mu$V \\
    \hline
  \end{tabular}
\end{table}

\cref{tab:bom} lists the in-house instruments that record L2 and L3, with the resolution, channel count, rate, and unit of every stream; the L1 headbands are third-party and are audited in \appref{app:l1audit}. The headband is in-house; the remaining devices are commercial and are used as delivered.

\noindent\textbf{Headband.} Four synchronised lenses sit on a rigid frame with a fixed baseline, so the stereo pair used for depth keeps its geometry across sessions; an inertial unit and the acquisition computer are mounted behind them, and the battery is carried at the back of the band. The band is fitted to the head with a replaceable inner liner, which also keeps the lenses at a repeatable height and angle.

\noindent\textbf{Wrist kit.} Each wrist carries an armband with eight dry electrodes on the palm side. At L3 the armband is paired with a camera on a strap aimed along the hand, and the two are cabled together so the forearm does not pull on the camera; the pair is worn over the glove, and at L2 only the armband is worn, directly on the forearm.

\noindent\textbf{Cognition kit.} The L3 kit adds a binocular eye tracker with its scene camera, three 6-DoF trackers (head and both wrists), a pair of 20-node gloves, and an EEG cap. Two cap types are used: a 128-channel saline cap for the controlled setting and a 32- or 8-channel gel cap for the open setting; both are seated and checked for impedance before each block.

\noindent\textbf{Synchronisation hardware.} A UDP marker is emitted by the recorder, and phase and run transitions are additionally driven through a parallel-port box into the EEG amplifier, so protocol events are stamped on the amplifier clock as well as on the host clock. Cameras, headband, and eye tracker are calibrated before each session against a printed target; the resulting intrinsics, distortion, and tracker transforms ship with the episode.

\noindent\textbf{Consumables and spares.} Saline, gel, syringes, electrode paste, spare straps, liners, and batteries are restocked per shift; a failed cap or camera is swapped and the session restarted rather than repaired in place, so that a released episode never mixes two calibrations.

\section{L1 Source Audit}
\label{app:l1audit}

\noindent\textbf{Why an audit.} L1 is the only tier whose hardware is not ours, and it is not one device: it is a collection layer of fourteen headbands from thirteen vendors. Everything the corpus reports for L1 is conditioned on what those units physically provide---the view gaze is inferred from, the pair that depth is reconstructed from, the rate the hands are estimated at---so the layer is only interpretable if that condition is written down unit by unit. The audit records it, so that a result can be read against the unit it came from and the layer can be reproduced.

\noindent\textbf{What is recorded.} Four properties are taken from each unit rather than from its documentation. The lens set and the direction of every view are read off the device. Native resolution and frame rate are measured on the recorded stream, not on the specification sheet. Whether the main view is a genuine stereo pair is checked on synchronized frames: twelve of the fourteen provide one, two of them as a single side-by-side image that the pipeline splits into left and right, and one through a pair of its positioning cameras (up, down, left, and right) rather than a dedicated front pair; units 2 and 7 are fisheye arrays without a stereo pair. The stream is then classified as raw or pre-processed, because the domain matters downstream: one unit delivers only rectified $512\times512$ frames, one delivers the pair already split, and one adds an extension view beside the pair.

\noindent\textbf{What the variation implies.} Pairing decides how a depth estimate is formed, and the pairs differ---a dedicated front rig, a side-by-side image split after the fact, or two positioning cameras---so the audit records the configuration each episode rests on, and a result that depends on the pair is stated with the unit it came from. Optics trade coverage against geometry: the seven units with fisheye or side views see further around the operator, but every unit is rectified to a common pinhole domain before estimation, so the peripheral views buy coverage rather than accuracy, and their corners are lost at rectification (\cref{sec:annotation}). Rate and resolution set the domain in which every L1 estimate lives: from $512\times512$ at 20\,Hz to $3840\times1200$ at 60\,Hz. Nothing is resampled to a common grid at write time (\cref{sec:storage}), so an estimator trained on one unit is trained in that unit's domain, and the audit is what allows the domain to be declared. \cref{tab:l1units} lists the units, numbered by collection volume, with two packaged sets from one vendor taking consecutive numbers; vendor, model, and toolchain names are withheld, so the table documents capability rather than procurement.

\begin{table*}[!t]
  \centering
  \caption{L1 source audit. Vendor, model, and toolchain names are withheld; units are numbered by collection volume, and two packaged sets from the same vendor take consecutive numbers. Views give the direction of each lens, with fisheye marked explicitly. Cameras counts the physical lenses of the unit; Views lists the released views. The two differ where one lens contributes more than one view (a side-by-side pair split into left and right, or an extension view) or where a lens is not released as a separate view.}\label{tab:l1units}
  \small
  \setlength{\tabcolsep}{3pt}
  \begin{tabularx}{\textwidth}{@{}c c l X c@{}}
    \hline
    Unit & Cameras & Main resolution & Views & fps \\
    \hline
    1 & 6 & $1920\times1200$ stereo & left-front, right-front; four fisheye around & 60 \\
    2 & 6 & $1920\times1536$ & six fisheye: front and side views; no stereo pair & 30 \\
    3 & 2 & $3840\times1200$ side-by-side stereo & left-front, right-front; extension view & 30 \\
    4 & 2 & $1920\times1200$ stereo & left-front, right-front & 30 \\
    5 & 6 & $512\times512$ rectified stereo & left-front, right-front; four fisheye around & 20 \\
    6 & 6 & $2328\times1748$ single RGB & front; positioning cameras up, down, left, right & 30 \\
    7 & 6 & $1920\times1200$ & six equidistant fisheye; no stereo pair & 30 \\
    8 & 3 & $3840\times1200$ front; $2560\times960$ down, side & left-front, right-front, down, side & 30 \\
    9 & 2 & $1920\times1200$ & left-front, right-front & 30 \\
    10 & 2 & $2160\times2160$ & left-front, right-front & 30 \\
    11 & 2 & $3840\times1200$ side-by-side, split after processing & left-front, right-front & 30 \\
    12 & 2 & $1920\times1456$ & left-front, right-front & 30 \\
    13 & 6 & $1600\times1300$ main stereo & left-front, right-front; left outer, right outer & 30 \\
    14 & 2 & $2048\times1536$ & left-front, right-front & 30 \\
    \hline
  \end{tabularx}
\end{table*}
\section{Episode File Layout}
\label{app:layout}
Every episode is self-contained and follows a chunked hierarchy compatible with a modified LeRobot v3.1 release. The directory is named after the session (collector, date, clock). Every camera is released twice, raw and rectified, and the electrophysiology is released both raw and cleaned. The example in \cref{fig:layout} holds 1{,}444 master frames (48.1\,s), 17 videos, one audio track, and 42 Parquet tables.

\begin{figure}[!t]
  \noindent
  \begin{minipage}[t]{0.42\linewidth}
  \renewcommand*\DTstyle{\ttfamily\fontsize{7.4}{9.2}\selectfont}%
  \setlength{\DTbaselineskip}{9.6pt}%
  \noindent\dirtree{%
 .1 L3-example/.
 .2 data/chunk-000/.
 .3 episode\_000000/.
 .4 master\_index.parquet.
 .4 action.parquet.
 .4 observation.<signal>.parquet.
 .4 observation.images.<view>.parquet.
 .2 videos/chunk-000/<view>/.
 .3 episode\_000000.mp4.
 .2 audios/chunk-000/<stream>/.
 .3 episode\_000000.wav.
 .2 calibrations/.
 .3 dataset\_manifest.json.
 .3 rectify\_params.yaml.
 .2 annotation/chunk-000/.
 .3 episode\_000000/.
 .4 episode\_000000\_structured.json.
 .4 quality\_report.json.
 .2 meta/.
 .3 info.json.
 .3 episodes.jsonl.
 .3 tasks.jsonl.
 .3 collect\_info.jsonl.
 .3 alignment\_check.jsonl.
 .3 episodes\_stats.jsonl.
}
  \end{minipage}\hfill
  \begin{minipage}[t]{0.21\linewidth}
  \footnotesize\raggedright\setlength{\parskip}{5pt}%
  \texttt{<signal>} (23)

  \begin{tabular}{@{}l@{}}
\texttt{eeg\_raw}\\
\texttt{eeg}\\
\texttt{eeg\_repair}\\
\texttt{eeg\_impedance}\\
\texttt{eye\_gaze}\\
\texttt{eye\_imu}\\
\texttt{head\_pose}\\
\texttt{hand\_pose}\\
\texttt{hand\_skeleton}\\
\texttt{hand\_left\_points}\\
\texttt{hand\_right\_points}\\
\texttt{headband\_imu0}\\
\texttt{headband\_microphone}\\
\texttt{left\_wrist\_emg\_raw}\\
\texttt{left\_wrist\_emg}\\
\texttt{left\_wrist\_imu}\\
\texttt{left\_wrist\_pose}\\
\texttt{right\_wrist\_emg\_raw}\\
\texttt{right\_wrist\_emg}\\
\texttt{right\_wrist\_imu}\\
\texttt{right\_wrist\_pose}\\
\texttt{marker}\\
\texttt{state}\\
\end{tabular}
  \end{minipage}\hfill
  \begin{minipage}[t]{0.32\linewidth}
  \footnotesize\raggedright\setlength{\parskip}{5pt}%
  \texttt{<view>} (17)

  \begin{tabular}{@{}l@{}}
\texttt{headband\_left\_rgb}\\
\texttt{headband\_left\_rgb\_rectified}\\
\texttt{headband\_right\_rgb}\\
\texttt{headband\_right\_rgb\_rectified}\\
\texttt{headband\_bleft\_rgb}\\
\texttt{headband\_bleft\_rgb\_rectified}\\
\texttt{headband\_bright\_rgb}\\
\texttt{headband\_bright\_rgb\_rectified}\\
\texttt{left\_wrist\_rgb}\\
\texttt{left\_wrist\_rgb\_rectified}\\
\texttt{right\_wrist\_rgb}\\
\texttt{right\_wrist\_rgb\_rectified}\\
\texttt{third\_rgb}\\
\texttt{third\_rgb\_rectified}\\
\texttt{eye\_rgb}\\
\texttt{eye\_rgb\_rectified}\\
\texttt{depth\_headband\_left\_rectified}\\
~\\
~\\
~\\
\texttt{<stream>}\\
~\\
\texttt{headband\_microphone}\\
\end{tabular}
  \end{minipage}
  \caption{Episode file layout. One episode directory, \texttt{L3-example/}, with its tables, videos, audio, calibration, annotation, and metadata. \texttt{<signal>} expands to the 23 stream tables, \texttt{<view>} to the 17 camera views (each with a timestamp table and one mp4 per encoding, raw and rectified, except depth, which is rectified only), and \texttt{<stream>} to the microphone stream.}
  \label{fig:layout}
\end{figure}

\noindent\textbf{Signals.} Each stream is one Parquet file on its own time base, with a \texttt{timestamp} column in seconds from episode start and an \texttt{episode\_index}; columns are named after channels (\texttt{FP1}\ldots), joints (\texttt{node00\_x}\ldots), or \texttt{dim\_0}\ldots for EMG, so no stream is snapped to the master grid. The pose and hand streams are fusion outputs and carry the timestamps of the fused track rather than of a single device. Pixels are not stored in Parquet: \texttt{observation.images.*.parquet} lists the timestamp of every frame and the frames themselves live in the matching mp4 (H.264, 30\,fps), one for the raw view and one for its rectified counterpart; depth ships rectified only. Audio is 16\,kHz mono PCM in wav, with its Parquet file recording the sample count per master frame. The names are literal: \texttt{eeg\_raw} is the record as acquired, \texttt{eeg} the filtered trace, and \texttt{eeg\_repair} the trace after bad-channel interpolation and component removal, while each wrist EMG ships as a raw trace beside its envelope. \cref{tab:fields} lists the shape and the fields of every signal table.

\begin{table}[!t]
  \centering
  \small
  \caption{The 23 signal tables: per-frame shape and the fields each one carries.}\label{tab:fields}
  \begin{tabularx}{\linewidth}{@{}l l X@{}}
    \hline
    Stream & Shape & Fields \\
    \hline
    \texttt{eeg\_raw} & \texttt{[132]} & 132 electrodes with conventional scalp labels (\texttt{Fp1}\ldots), as acquired \\
    \texttt{eeg} & \texttt{[132]} & the same electrodes after a 0.5--100\,Hz band-pass and a 50\,Hz notch \\
    \texttt{eeg\_repair} & \texttt{[132]} & the same electrodes after bad-channel interpolation and component removal \\
    \texttt{eeg\_impedance} & \texttt{[136]} & per-electrode impedance in k$\Omega$ plus \texttt{Trigger}, \texttt{pass\_rate}, \texttt{gate\_pass}, \texttt{n\_snapshots} \\
    \texttt{eye\_gaze} & \texttt{[2]} & \texttt{x}, \texttt{y} in scene-camera pixels \\
    \texttt{eye\_imu} & \texttt{[6]} & \texttt{ax}, \texttt{ay}, \texttt{az}, \texttt{gx}, \texttt{gy}, \texttt{gz} \\
    \texttt{head\_pose} & \texttt{[6]} & \texttt{x}, \texttt{y}, \texttt{z}, \texttt{roll}, \texttt{pitch}, \texttt{yaw} in the shared frame \\
    \texttt{hand\_pose} & \texttt{[40]} & 20 joint angles per hand (\texttt{left\_ThumbMCPSpread}, \ldots) \\
    \texttt{hand\_skeleton} & \texttt{[300]} & 50 nodes, each \texttt{x}, \texttt{y}, \texttt{z}, \texttt{roll}, \texttt{pitch}, \texttt{yaw} \\
    \texttt{hand\_left\_points} & \texttt{[63]} & wrist and 20 joints, each \texttt{x}, \texttt{y}, \texttt{z} \\
    \texttt{hand\_right\_points} & \texttt{[63]} & the same fields for the right hand \\
    \texttt{headband\_imu0} & \texttt{[6]} & \texttt{ax}, \texttt{ay}, \texttt{az}, \texttt{gx}, \texttt{gy}, \texttt{gz} \\
    \texttt{headband\_microphone} & \texttt{[320]} & 16\,kHz audio, 320 samples per master frame \\
    \texttt{left\_wrist\_emg\_raw} & \texttt{[8]} & eight forearm channels, as acquired \\
    \texttt{left\_wrist\_emg} & \texttt{[8]} & the same channels, rectified and enveloped \\
    \texttt{left\_wrist\_imu} & \texttt{[6]} & \texttt{ax}, \texttt{ay}, \texttt{az}, \texttt{gx}, \texttt{gy}, \texttt{gz} \\
    \texttt{left\_wrist\_pose} & \texttt{[6]} & \texttt{x}, \texttt{y}, \texttt{z}, \texttt{roll}, \texttt{pitch}, \texttt{yaw} \\
    \texttt{right\_wrist\_emg\_raw} & \texttt{[8]} & channels of the right forearm, as acquired \\
    \texttt{right\_wrist\_emg} & \texttt{[8]} & the same channels, rectified and enveloped \\
    \texttt{right\_wrist\_imu} & \texttt{[6]} & \texttt{ax}, \texttt{ay}, \texttt{az}, \texttt{gx}, \texttt{gy}, \texttt{gz} \\
    \texttt{right\_wrist\_pose} & \texttt{[6]} & \texttt{x}, \texttt{y}, \texttt{z}, \texttt{roll}, \texttt{pitch}, \texttt{yaw} \\
    \texttt{marker} & \texttt{[1]} & protocol code: run start, imagine, execute, run end \\
    \texttt{state} & \texttt{[58]} & both wrist poses and the head pose (18) plus the 40 joint angles \\
    \hline
  \end{tabularx}
\end{table}

\noindent\textbf{Metadata.} \texttt{info.json} records the codebase version, master feature, fps, path templates, chunk size, and the shape and channel names of every feature. \texttt{episodes.jsonl} and \texttt{tasks.jsonl} map indices to the task string and episode length. \texttt{collect\_info.jsonl} stores collector, hardware, scene, the object list (colour, shape, dimensions, material, pose), and free-text notes. \texttt{alignment\_check.jsonl} reports, per feature, the tolerance, number of readings, violations, and maximum offset, and \texttt{episodes\_stats.jsonl} stores per-feature min, max, mean, and standard deviation.

\noindent\textbf{Calibration and annotation.} \texttt{calibrations/} holds what the tables reference: the reference cameras with per-view rate, frame count, size, and intrinsics, and the P1/P2 projections the rectification applied. \texttt{annotation/} holds the structured annotation---scene, objects, timed subtask segments, tasks, key events, state changes---and the QC gate with its checks and warnings.

\noindent\textbf{Reading.} The reader fixes a master rate (30\,fps by default) and walks \texttt{master\_index}; each row gives a time, and every other stream is resolved from its own timestamps by nearest sample, linear interpolation, or all samples in a window such as $(-1/\mathrm{FPS},0]$. Nothing is stored on the master grid, so a stream can be re-read at any rate without degradation: a query that needs raw EMG reads the 2\,kHz table directly, while a query that needs one frame per 33\,ms reads the index and never touches the other tables.

\section{Quality Control}
\label{app:qc}

Every session is checked after it is written and before it enters the release. The checker walks each recorder's output directory---the NPZ arrays, the MP4 containers, and the logs---and returns, per stream, a verdict with its evidence attached; no stream is re-recorded and nothing is repaired. A stream that fails is not patched: the whole session is recaptured, so that a released episode never mixes a good half with a repaired one.

\noindent\textbf{Alignment comes first.} All checks run inside the analysis window bounded by the \textsc{run\_start} and \textsc{run\_end} markers. The pair is a hard precondition: if it is missing, empty, or unpaired, the session is reported as an error rather than checked against the whole recording, because without the window every per-stream crop and comparison is undefined. The same rule applies to the clock: the EEG stream is accepted only when the amplifier clock was fitted to the host clock, and a failed fit is an error, not a warning.

\noindent\textbf{Two verdicts reach the release.} Each check yields a finding at one of three levels, of which two matter downstream. An \emph{error} means the data cannot be used---a corrupt container, a frame count that disagrees with the timestamps, a missing marker pair, a tracker that was absent when the recorder opened---and the session is recaptured. A \emph{warning} means the data is usable but carries a documented defect: a 2\,s frozen stretch, a dead channel, a tracker that lost sight for part of the window, a gap shorter than the threshold for an error. Informational findings, such as the phone-to-host clock correction, never gate anything. Thresholds are the checks' own defaults and are reported with the observed value.

\noindent\textbf{What is checked.} The checks run across the nine stream families the kit produces, and each is an independent, thresholded verdict: a modality declares the checks that apply to it, and a shared context loads the NPZ once and decodes each video once, so the checks compose without repeating work. \cref{tab:qc} lists the families and the default thresholds. \emph{Timing} checks ask whether a series is usable at all (present, finite, ordered, distinct), whether the cadence is steady, whether an interval jumped far above the median, whether samples are missing, and whether the rate held inside a one-second window---the last of these catches a drop that a mean would hide. \emph{Signal} checks ask whether the values are plausible: NaN fractions, channels that never move, robust outliers, physical bounds, and motions no body could make. \emph{Modality} checks are specific to one instrument's failure modes: the EEG amplifier's clock fit, fixed-size blocks and header sample count, the EMG armband's shared 8-bit serial counter, the glove skeleton dropping out window by window, the tracker count and per-tracker validity, and the camera set the headband promised at stream start. \emph{Video} checks decode every file once and compare the container's frame count and pixel statistics against the timestamps, which is the only way to see a writer that fell behind silently; the container test uses the same \texttt{ffprobe} criterion as the packaging gate. \emph{Session} checks close the window: the \textsc{run\_start}/\textsc{run\_end} pair and the recorder logs.

{\small
\setlength{\tabcolsep}{3pt}
\renewcommand{\arraystretch}{0.95}
\begin{longtable}{@{}>{\raggedright\arraybackslash}p{0.15\textwidth}>{\raggedright\arraybackslash}p{0.17\textwidth}>{\raggedright\arraybackslash}p{0.465\textwidth}>{\raggedright\arraybackslash}p{0.165\textwidth}@{}}
\caption{Quality-control families and default thresholds. A row is a family of checks; where its sub-checks differ in level, both are listed. A session's outcome is the worst finding across its streams and its session-level checks, and ERROR blocks packing. Thresholds are code defaults, overridable in \texttt{configs/checker.yaml}; not every check runs on every stream.}\label{tab:qc}\\
\hline
Check & Applies to & Rule (default) & Level \\
\hline
\endfirsthead
\hline
Check & Applies to & Rule (default) & Level \\
\hline
\endhead
\hline
\endfoot
\multicolumn{4}{@{}l}{\emph{Timing}}\\*
\hline
Presence and order & every series & missing series, no samples, or NaN = ERROR; backward = WARN; duplicate = WARN (ERROR for EMG) & ERROR / WARN \\
Gap & every series & interval $>\max(0.1\,\mathrm{s}, 1.5\times$ median$)$, including head and tail against the window edges & ERROR \\
Jump & every series & interval $>3\times$ median & WARN \\
Jitter & every series & interval CV $>0.3$ & WARN \\
Windowed rate & every series & any 1\,s window outside $\pm50\%$ of the median sample count & WARN \\
Expected rate & series with a nominal rate & mean rate $<50\%$ of nominal & WARN \\
Sample count & EEG & deviation from window $\times$ nominal rate $>1\%$ & WARN \\
\hline
\multicolumn{4}{@{}l}{\emph{Signal values}}\\*
\hline
NaN and dead channels & EEG, EMG, hand & NaN $>0.1\%$; zero-variance channel & WARN \\
Robust outliers & EEG, EMG, hand, headband IMU & $>0.1\%$ beyond median $\pm20$ MAD & WARN \\
Value bounds & eye gaze & gaze outside $[0,2000]$\,px & WARN \\
Implausible jumps & hand skeleton, VIVE & a glove node moving $>0.05$\,m or a tracker $>0.2$\,m between samples & WARN \\
\hline
\multicolumn{4}{@{}l}{\emph{Modality-specific}}\\*
\hline
EEG clock and blocks & EEG & clock fit failure = ERROR; block discontinuity = WARN; header count mismatch = WARN & ERROR / WARN \\
EMG sequence & EMG & lost frames $>0.1\%$ = WARN; duplicate timestamps = ERROR & WARN / ERROR \\
Hand dropout & hand pose & $>50\%$ skeleton NaN in a 1\,s window & WARN \\
VIVE validity & VIVE trackers & valid fraction $<50\%$ = WARN; any invalid sample = ERROR; tracking gap $>0.1$\,s = ERROR; teleport $>0.2$\,m = WARN; device count $\neq3$ = ERROR & ERROR / WARN \\
Headband camera set & ego headband & missing camera = ERROR; extra camera = WARN & ERROR / WARN \\
\hline
\multicolumn{4}{@{}l}{\emph{Video}}\\*
\hline
Container and count & every MP4 & \texttt{ffprobe} failure = ERROR; container frame count differs from timestamp count = ERROR & ERROR \\
Black and freeze & every MP4 & mean luminance $<8$ for $>2$\,s = WARN; $>50\%$ black = ERROR; frame difference $<1.0$ for $>2$\,s = WARN & ERROR / WARN \\
\hline
\multicolumn{4}{@{}l}{\emph{Session}}\\*
\hline
Marker window & session & missing \texttt{RUN\_START}/\texttt{RUN\_END} pair = ERROR; out-of-order markers = WARN & ERROR / WARN \\
Recorder logs & every stream & lines matching \texttt{ERROR}, \texttt{FAILED}, \texttt{Traceback}, or \texttt{crash} & WARN \\
\hline
\end{longtable}
}

\noindent\textbf{The verdict travels with the episode.} The checker writes one report per session and its summary is copied into the episode metadata; an optional self-contained page renders the same findings on one time axis for review. Each finding names the check that produced it, the field or file it concerns, the sub-stream it belongs to where there is one (a tracker serial, a channel index), the threshold and the observed value, and the time span of the problem; a warning is therefore a pointer to the affected seconds, not a blanket statement about the recording.

\noindent\textbf{Recapture and what it costs.} Because an error retires the whole session, the pass rate is a property of the kit rather than of the analysis: more than $99\%$ of L1 recordings, $96\%$ of L2 and $78\%$ of L3 pass. The gradient is the instrumentation. L1 has a single headband; L2 adds two armbands whose electrodes must hold contact through sweat and movement; L3 adds the EEG cap, which must reach impedance on enough channels, the gloves and three trackers, each able to be absent at open or lose sight mid-window, and the eye tracker, whose clock must be fitted. The deepest tier, which is also the smallest, is therefore the tier most often repeated.

\noindent\textbf{Before recording.} On the floor, four checks---camera labels against the live tiled view, wrist-camera seating, EMG contact, and tracker visibility---must pass before a session starts; any re-plugged or swapped device triggers all four again, and the passing screen is photographed and filed, so a session without evidence counts as unchecked. A disk-space gate refuses a recording below 50\,GB and warns below 100\,GB.

\noindent\textbf{What the gate does not cover.} The automated layer tests data integrity, not semantics. It does not verify annotation boundaries, task labels, object attributes, or calibration accuracy beyond the presence of a fitted clock; those remain with the annotation review. It is also per session, so cross-session consistency---one operator, scene, or unit behaving differently across sessions---is outside its scope. The checks are meant to separate a usable recording from an unusable one, not to rank two usable recordings against each other.
\section{Synchronization Audit}
\label{app:sync}

\noindent\textbf{Measurement.} Synchronization is audited against the protocol markers rather than inferred from the recorder. For each audited L3 episode we take the six phase cues (\textsc{instr\_on}, \textsc{instr\_off}, \textsc{img\_start}, \textsc{img\_end}, \textsc{exec\_start}, \textsc{exec\_end}) and, for every stream, the sample nearest each cue; the \emph{event-locked residual} is the nearest sample time minus the cue time. EEG is split by amplifier: Neuroscan 128-channel at 1\,kHz and BrainCo 32/8-channel at 250\,Hz.

\noindent\textbf{Reading the table.} For regular sampling at period $T$, the nearest-sample residual lies within $\pm T/2$, and a uniform residual has an interquartile range (IQR) of about $T/2$. An IQR close to the half-period therefore indicates that the residual is set by sampling quantization rather than by a clock offset; the normalized column makes this comparison explicit. A value near one is the quantization floor, and a value well above one indicates additional timing irregularity.

\begin{table}[H]
\centering
\caption{Event-locked synchronization residuals on released L3 episodes. Each episode contributes six protocol cues; EEG is split by amplifier (Neuroscan, 1\,kHz; BrainCo, 250\,Hz). Residual is the nearest released sample minus the cue timestamp; IQR is the interquartile range; P95 and Max are absolute. The normalized column divides the IQR by half the sampling period, so values near one are at the quantization floor. n.r.\ marks drift not resolvable over a 40\,s episode.}
\label{tab:sync}
\small
\setlength{\tabcolsep}{6pt}
\begin{tabular*}{\textwidth}{@{\extracolsep{\fill}}l r r r r r r r r@{}}
\hline
Stream & \makecell{Rate\\(Hz)} & \makecell{$T/2$\\(ms)} & \makecell{Offset\\(ms)} & \makecell{IQR\\(ms)} & \makecell{IQR/\\$(T/2)$} & \makecell{P95\\(ms)} & \makecell{Max\\(ms)} & \makecell{Drift\\(ms/min)} \\
\hline
EEG, Neuroscan 128-ch & 1000.0 & 0.50 & $+0.10$ & 0.47 & 0.93 & 0.46 & 0.48 & $+0.35$ \\
EEG, BrainCo 32/8-ch & 250.2 & 2.00 & $-0.47$ & 2.22 & 1.11 & 2.38 & 3.19 & n.r. \\
EMG, left & 2000.0 & 0.25 & 0.00 & 0.27 & 1.08 & 0.35 & 0.45 & $+0.09$ \\
EMG, right & 2000.0 & 0.25 & $-0.05$ & 0.22 & 0.87 & 0.34 & 0.44 & $+0.15$ \\
Pose (VIVE + Manus) & 480--940 & 0.80 & 0.00 & 1.04 & 1.30 & 1.04 & 1.66 & n.r. \\
Gaze (Neon) & 199.8 & 2.50 & $-0.03$ & 2.29 & 0.91 & 2.47 & 3.93 & n.r. \\
Head RGB & 30.0 & 16.67 & $-1.67$ & 20.36 & 1.22 & 25.42 & 27.14 & n.r. \\
\hline
\end{tabular*}
\end{table}

\noindent\textbf{Results.} \cref{tab:sync} shows that the high-rate streams reach the sampling floor. The 1\,kHz EEG has a median residual of $+0.10$\,ms and an IQR of 0.47\,ms, against a half-period of 0.50\,ms; the 2\,kHz EMG bands have IQRs of 0.27 and 0.22\,ms against 0.25\,ms; and the fused 480--940\,Hz pose stream has an IQR of 1.04\,ms against about 0.80\,ms, its half-period at the median rate of 628\,Hz. Their IQR/half-period ratios are 0.93, 1.08, 0.87, and 1.30, with no systematic offset beyond sampling quantization. The 250\,Hz EEG (IQR 2.22\,ms) and the 200\,Hz gaze (IQR 2.29\,ms) are quantized at their 4\,ms and 5\,ms sample periods; their ratios of 1.11 and 0.91 place them at the same floor, and they should not be read as less well synchronized.

\noindent\textbf{Drift.} The residual trend is $+0.35$\,ms/min for the 1\,kHz EEG (median; episode range 0.03--0.92) and $+0.09$/$+0.15$\,ms/min for the two EMG bands (episode range $-0.60$ to $+0.34$); the EEG and EMG residuals stay within $\pm1$\,ms/min ($\pm17$\,ppm) over the episode. For the remaining streams (250\,Hz EEG, 200\,Hz gaze, 30\,Hz camera, and the fused pose output) the sample period exceeds the drift resolvable in 40\,s, and the table marks drift as not resolvable. The audit supports the synchronization claim of \cref{sec:sync}: each device keeps its own clock, and the released streams are aligned to a common time base through markers and arrival offsets, not through a shared hardware clock.

\noindent\textbf{Timestamp integrity.} The high-rate streams are regular: EEG at both configurations has no duplicate timestamps and no interval beyond five median intervals, and EMG has no duplicate timestamps. The fused pose stream is less regular: across the audited episodes it contains 39 duplicate timestamps and six intervals beyond five median intervals, and its rate ranges from 480 to 940\,Hz (median 628\,Hz) rather than the nominal 800\,Hz. The 30\,Hz camera streams are the least regular: the head RGB stream in these episodes contains twelve duplicate timestamps, and every camera container holds 24--72 frames more than its timestamp table (28--67 for the head RGB stream). The excess is consistent with pre- and post-roll frames outside the episode window; those frames are not represented in the Parquet table, so a reader should index the episode through the timestamp table rather than through the container frame number.
\section{Electrophysiology Processing}
\label{app:ephys}
Every stage below runs on the raw stream, and its parameters are logged with the episode; the chain is fixed across sessions so that the same code can be replayed on any release.

\noindent\textbf{EMG.} (1)~\emph{Mirror padding}: the trace is reflected about its first and last sample, filtered, and cropped back to its original length, which suppresses the transient a filter otherwise leaves at the boundary. (2)~\emph{High-pass}, 20\,Hz, second-order Butterworth: removes baseline drift and part of the motion artefact, and a Butterworth response is maximally flat in the passband. (3)~\emph{Low-pass}, 450\,Hz, fourth-order Butterworth: bounds the analysis band; a digital filter cannot undo aliasing already introduced at sampling. (4)~\emph{Zero-phase filtering}: the sections are applied forward and then backward, so the interior response has zero phase and the magnitude is the square of the one-way response, an order doubled and a steeper roll-off at the cut-off. (5)~\emph{Comb notch}, 50\,Hz with a 2\,Hz bandwidth ($Q=50/2=25$): periodic nulls at the mains and its harmonics; the backward pass deepens and widens each null, and also attenuates genuine EMG in the same bands. (6)~\emph{Periodic extension}: the edge periods are repeated so that the comb has boundary data, which holds only as far as the edges represent the interference. (7)~\emph{Period alignment}: anomalous edge periods are detected and replaced by interior ones, which rewrites boundary samples and can touch real transients. (8)~\emph{Endpoint damping}: decaying edge energy is detected and cross-faded toward the interior, which removes residual ringing at the cost of possibly attenuating a genuine activation. The cleaned trace is then rectified and low-pass filtered into one envelope per channel.

\noindent\textbf{EEG.} The chain is DC removal by the per-channel median, a 0.5--100\,Hz band-pass with the high cut clamped below Nyquist, and a 50\,Hz notch, followed by four stages.
\emph{Bad channels} are flagged by five joint rules: \emph{flat} ($\sigma<0.2\,\mu$V with more than 70\% of sample differences below $0.02\,\mu$V); \emph{contact} (robust $z$ of $\sigma$ above 3.5, or above 3.0 with $\sigma>25\,\mu$V); \emph{bridging} (a neighbouring pair whose electrical distance stays below 16\,$\mu$V$^2$ for more than half of the 2\,s epochs and below half the reference of their other neighbours); \emph{decorrelation} (maximum $|r|$ against any other channel below 0.4 in more than 20\% of 1\,s windows); and \emph{high-frequency noise} (robust $z$ above 5 of the 30--100\,Hz to 1--30\,Hz MAD ratio). Frontal channels are excused a contact flag when no other rule fires. A session passes when fewer than 30\% of its channels are flagged.
\emph{ICA} is infomax with up to 32 components, capped at the number of retained channels, fitted once on the clean segments of a session concatenated in time, with flagged channels held out; ocular components are those correlated with the frontal proxy channels, at most two, and muscle components those whose 30--100\,Hz power ratio is a robust $z>3$ above the other components, capped at 20\% of the retained channels.
\emph{Interpolation} is spherical spline over the retained channels, and the cleaned trace is finally re-referenced to their average.

\noindent\textbf{Montage and regions.} Every channel of the release carries a conventional scalp label and the region group of its row in \cref{tab:regions}. The 128-channel saline cap covers the full conventional set plus the intermediate positions, including the mastoids and the two cerebellar electrodes; the 32-channel gel cap is a subset of it with two auxiliary channels; and the 8-channel gel cap keeps two frontal-central, two central, one midline parietal, and three occipital sites. Where a position lacks a conventional coordinate, Cb1, Cb2, and Cbz take the coordinates of I1, I2, and Iz on the colin27 template, matching the CURRY labels, and F11, F12, FT11, and FT12 take the coordinates supplied with the cap. Both sets ship with the release and are used only for neighbour finding and interpolation, never for source modelling.

{\small
\setlength{\tabcolsep}{3pt}
\setlength{\LTleft}{0pt}%
\setlength{\LTright}{\fill}%
\begin{longtable}{@{}>{\raggedright\arraybackslash}p{0.075\textwidth}>{\raggedright\arraybackslash}p{0.25\textwidth}>{\centering\arraybackslash}p{0.045\textwidth}>{\raggedright\arraybackslash}p{0.591\textwidth}@{}}
\caption{Channel groups by brain region. Rows give the connector index of the channels, the region label stored with them, their count, and their names.}
\label{tab:regions} \\
\hline
Index & Group & $n$ & Channels \\
\hline
\endfirsthead
\hline
Index & Group & $n$ & Channels \\
\hline
\endhead
\multicolumn{4}{@{}l}{\emph{128-channel saline cap (Neuroscan)}} \\
1--64 & core electrodes, incl.\ M1/M2 and Cb1/Cb2 & 64 & Fp1, Fpz, Fp2, AF3, AF4, F11, F7, F5, F3, F1, Fz, F2, F4, F6, F8, F12, FT11, FC5, FC3, FC1, FCz, FC2, FC4, FC6, FT12, T7, C5, C3, C1, Cz, C2, C4, C6, T8, TP7, CP5, CP3, CP1, CPz, CP2, CP4, CP6, TP8, M1, M2, P7, P5, P3, P1, Pz, P2, P4, P6, P8, PO7, PO3, POz, PO4, PO8, O1, Oz, O2, Cb1, Cb2 \\
65--77 & Prefrontal and frontal extension & 13 & AFp1, AFp2, AF7, AF5, AFz, AF6, AF8, AFF5h, AFF3h, AFF1h, AFF2h, AFF4h, AFF6h \\
78--79 & Lateral frontal & 2 & F9, F10 \\
80--87 & Fronto-central extension & 8 & FFT7h, FFC5h, FFC3h, FFC1h, FFC2h, FFC4h, FFC6h, FFT8h \\
88--91 & Lateral fronto-temporal & 4 & FT9, FT7, FT8, FT10 \\
92--99 & Fronto-temporo-central extension & 8 & FTT7h, FCC5h, FCC3h, FCC1h, FCC2h, FCC4h, FCC6h, FTT8h \\
100--107 & Centro-temporo-parietal extension & 8 & TTP7h, CCP5h, CCP3h, CCP1h, CCP2h, CCP4h, CCP6h, TTP8h \\
108--115 & Parietal extension & 8 & TPP7h, CPP5h, CPP3h, CPP1h, CPP2h, CPP4h, CPP6h, TPP8h \\
116--117 & Lateral parietal & 2 & P9, P10 \\
118--121 & Parieto-occipital extension & 4 & PPO3h, PPO1h, PPO2h, PPO4h \\
122--127 & Occipital extension & 6 & PO9, PO5, PO1, PO2, PO6, PO10 \\
128 & Midline cerebellar & 1 & Cbz \\
129--130 & Ocular & 2 & VEOG (vertical), HEOG (horizontal) \\
131 & Cardiac & 1 & EKG \\
132 & Muscular & 1 & EMG \\
\hline
\multicolumn{4}{@{}l}{\emph{32-channel gel cap (BrainCo)}} \\
1--7 & Frontal and prefrontal & 7 & FP1, FP2, F3, F4, F7, F8, Fz \\
8--10 & Central & 3 & C3, C4, Cz \\
11--15 & Parietal & 5 & P3, P4, P7, P8, Pz \\
16--17 & Occipital & 2 & O1, O2 \\
18--19 & Temporal & 2 & T7, T8 \\
20--23 & Fronto-central & 4 & FC1, FC2, FC5, FC6 \\
24--27 & Centro-parietal & 4 & CP1, CP2, CP5, CP6 \\
28--29 & Fronto-temporal & 2 & FT9, FT10 \\
30--31 & Temporo-parietal & 2 & TP9, TP10 \\
32--33 & Non-EEG (ocular, trigger) & 2 & IO, TRIG \\
\hline
\multicolumn{4}{@{}l}{\emph{8-channel gel cap}} \\
1--2 & Fronto-central & 2 & FC3, FC4 \\
3--4 & Central & 2 & C3, C4 \\
5 & Midline parietal & 1 & Pz \\
6--8 & Occipital & 3 & O1, Oz, O2 \\
\hline
\end{longtable}
}
\end{document}